\documentclass[journal,hideappendix]{vgtc}                     %
\vgtccategory{Research}

\vgtcpapertype{algorithm/technique}

\title{Dream360: Diverse and Immersive Outdoor Virtual Scene Creation via Transformer-Based 360$^\circ$ Image Outpainting}

\author{%
  \authororcid{Hao Ai}{0000-0003-2104-3352}, Zidong Cao, Haonan Lu, Chen Chen, Jian Ma, Pengyuan Zhou, Tae-Kyun Kim, Pan Hui, and Lin Wang$^\dagger$
}

\authorfooter{
  $^\dagger$ Corresponding author \\
  \item
  	H. Ai, Z. Cao are with HKUST(GZ), Guangzhou, China. E-mail: \{hai033\,$|$\,zcao740\}@connect.hkust-gz.edu.cn\,.
  \item
  	H.Lu, C. Chen, J. Ma are with OPPO, Shenzhen, China. E-mail: \{luhaonan\,$|$\,chenchen4\,$|$\,majian2\}@oppo.com\,.

  \item 
    P. Zhou is with USTC, Hefei, China. E-mail: pyzhou@ustc.edu.cn\,.

  \item 
    T-K. Kim is with KAIST, Daejeon, Korean, and ICL, London, United Kingdom. E-mail: kimtaekyun@kaist.ac.kr.
    
  \item 
    P. Hui and L. Wang are with HKUST(GZ), Guangzhou, and HKUST, Hong Kong SAR, China. E-mail: \{panhui\,$|$\,linwang\}@ust.hk\,.
    
}

\abstract{%
360$^\circ$ images, with a field-of-view (FoV) of $180^\circ \times 360^\circ$, provide immersive and realistic environments for emerging virtual reality (VR) applications, such as virtual tourism, where users desire to create diverse panoramic scenes from a narrow FoV photo they take from a viewpoint via portable devices. It thus brings us to a technical challenge: `\textit{How to allow the users to freely create diverse and immersive virtual scenes from a narrow FoV image with a specified viewport}?'
To this end, we propose a transformer-based 360$^\circ$ image outpainting framework called \textbf{Dream360}, which can generate diverse, high-fidelity, and high-resolution panoramas from user-selected viewports, considering the spherical properties of 360$^\circ$ images. Compared with existing methods,~\eg, \cite{Akimoto2022DiverseP3}, which primarily focus on inputs with rectangular masks and central locations while overlooking the spherical property of 360$^\circ$ images, our Dream360 offers higher outpainting flexibility and fidelity based on the spherical representation. Dream360 comprises two key learning stages: \textbf{(I)} codebook-based panorama outpainting via Spherical-VQGAN (S-VQGAN), and \textbf{(II)} frequency-aware refinement with a novel frequency-aware consistency loss. 
Specifically, S-VQGAN learns a sphere-specific codebook from spherical harmonic (SH) values, providing a better representation of spherical data distribution for scene modeling.
The frequency-aware refinement matches the resolution and further
improves the semantic consistency and visual fidelity of the generated results. 
Our Dream360 achieves significantly lower Frechet Inception Distance (FID) scores and better visual fidelity than existing methods. We also conducted a user study involving 15 participants to interactively evaluate the quality of the generated results in VR, demonstrating the flexibility and superiority of our Dream360 framework.
}


\teaser{
  \centering
  \includegraphics[width=0.9\linewidth]{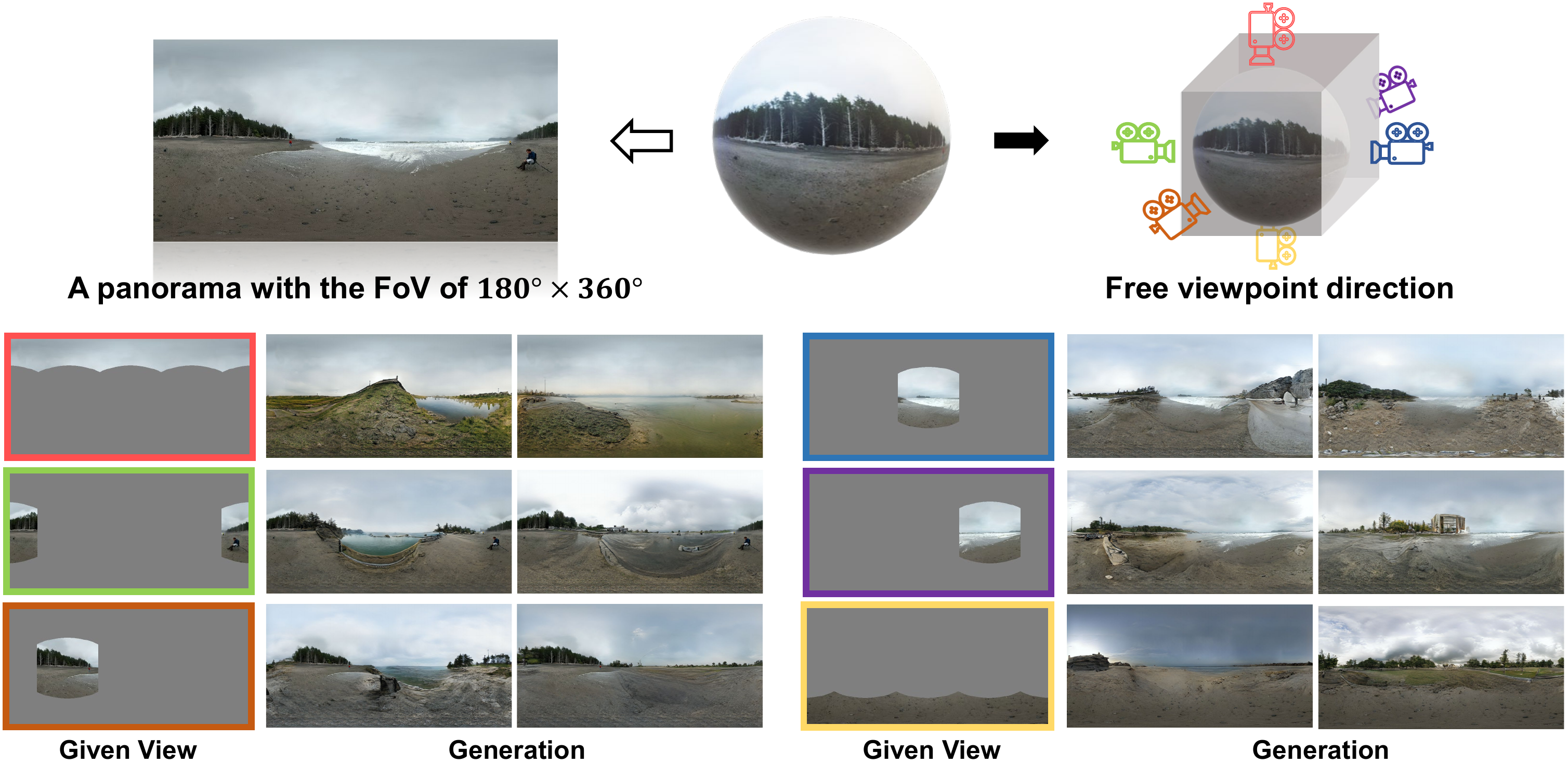}
    \vspace{-5pt}
  \caption{Top: Different viewpoints of a panorama; Bottom: Diverse outdoor virtual scene creation results from different viewpoints. 1st to 3rd rows: Left: \textcolor{red}{ceiling}, \textcolor{green}{left}, \textcolor{brown}{front}; Right: \textcolor{blue}{right}, \deeppurple{back}, \textcolor{yellow}{floor}.}
\vspace{-5pt}
  \label{fig:coverfig}
}

\renewcommand{\manuscriptnotetxt}{Manuscript revised xxx}

\graphicspath{{figs/}} 

\usepackage{tabu}                      
\usepackage{booktabs}                  
\usepackage{lipsum}                    
\usepackage{mwe}                       
\usepackage{microtype} 
\usepackage{mathptmx}                  
\usepackage{cite}
\usepackage{tabu}
\usepackage{multirow}
\usepackage{subfigure}
\usepackage{colortbl}
\usepackage [svgnames]{xcolor}
\definecolor{deeppurple}{rgb}{0.4, 0, 1}
\newcommand{\deeppurple}[1]{\color{deeppurple}{#1}}
\usepackage{amsmath}
\usepackage{amssymb}
\usepackage[normalem]{ulem}

\newcommand{\etal}{\textit{et al}.}

\newcommand{\ie}{\textit{i}.\textit{e}.}
\newcommand{\eg}{\textit{e}.\textit{g}.}

\newcommand{\RN}[1]{%
  \textup{\uppercase\expandafter{\romannumeral#1}}%
}
\usepackage{enumitem}
\setenumerate[1]{itemsep=0pt,partopsep=0pt,parsep=\parskip,topsep=5pt}
\setitemize[1]{itemsep=0pt,partopsep=0pt,parsep=\parskip,topsep=5pt}
\setdescription{itemsep=0pt,partopsep=0pt,parsep=\parskip,topsep=5pt}
\begin{document}


\firstsection{Introduction}
\label{intro}
\maketitle

The ability to observe complete surroundings in one shot and choose any viewport for immersive experience (see Fig.\ref{fig:coverfig}(a)) has led to the growing popularity of 360$^\circ$ images (\textit{a.k.a.} panoramas)\footnote{With equirectangular projection (ERP) images by default.} flourishing virtual reality (VR) applications~\cite{Ardouin2014StereoscopicRO, Martin2022ScanGAN360AG,Marrinan2021RealTimeOS,Vermast2023Introducing3T,Arora2022AugmentingIT,Li2022BulletCF,Wang2020Transitioning360CN}, such as virtual tourism. 
In this case, 
users wish to create diverse virtual panoramic scenes from a narrow FoV (NFoV) photo they take from a viewpoint via portable devices (\eg, smartphones). This way, it offers the users personalized contents and enables them to perceive and interact with virtual environments. This motivates us to study diverse high-fidelity and high-resolution panorama generation while users provide NFoV images at desired viewpoints. 

Several methods~\cite{Akimoto2019360DegreeIC, Hara2021SphericalIG, Akimoto2022DiverseP3,Han2020PIINETA3} have leveraged the generative models, such as variational auto-encoders (VAEs)\cite{Kingma2014AutoEncodingVB} and generative adversarial networks (GANs)\cite{Goodfellow2014GenerativeAN}, to synthesize panoramas by predicting surrounding pixels from cropped central regions. However, their outputs are deterministic in nature because the models are trained in a supervised manner using single instance labels.
Recently,~\cite{Akimoto2022DiverseP3} achieves more diverse panorama outpainting based on the planar image-generative model, Taming Transformer (TT)~\cite{Esser2021TamingTF}, which applies an auto-regressive transformer to model the scene using the codes provided by VQGAN~\cite{Esser2021TamingTF} and employs an extra adjustment stage to enhance the outpainting performance. 
Furthermore, the WS-perceptual loss~\cite{Sun2017WeightedtoSphericallyUniformQE} is applied to focus on information-rich regions around the sphere equator. 

These methods, however, suffer from the following problems: 1) The input is limited to a rectangular mask and a central location, lacking viewpoint flexibility; 2) The codebook learned via vanilla VQGAN ignores the spherical property and shows severe artifacts on generated panoramas; 3) As its adjustment stage only matches the given input and generated regions in the RGB domain, it is difficult to improve visual fidelity and restore structural details because of the \textit{spectral bias}, the learning bias of neural networks towards low-frequency functions highlighted in~\cite{Rahaman2018OnTS,TancikSMFRSRBN20}. Therefore, existing methods can hardly generate diverse high-fidelity panoramas from a freely given view direction.

\begin{figure*}[ht]
  \centering
  \includegraphics[width=0.9\linewidth]{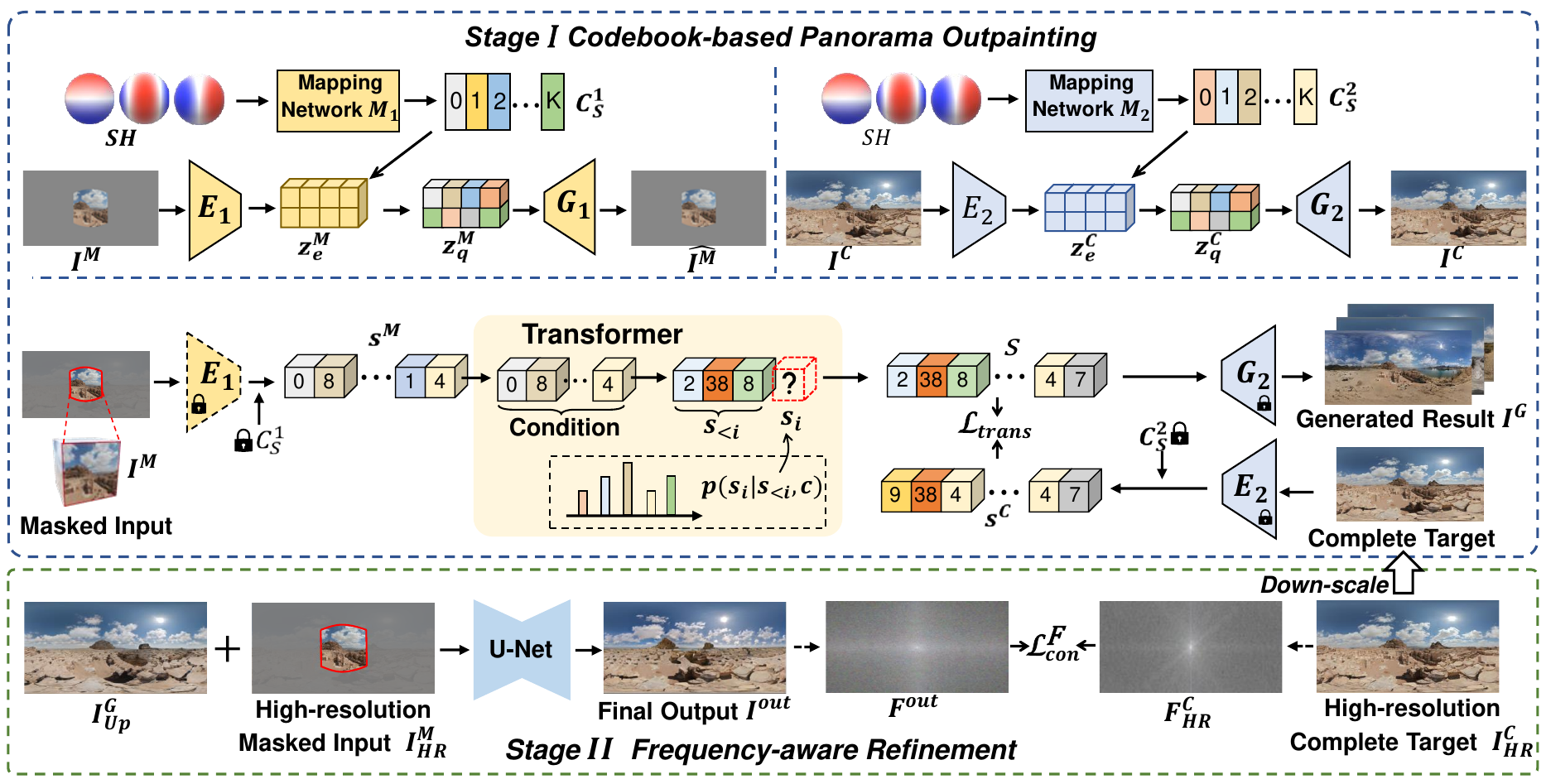}
  \vspace{-5pt}
  \caption{Overview of Dream360, consisting of two stages: Codebook-based panorama outpainting and Frequency-aware refinement.}
  \vspace{-12pt}
 \label{fig:overview}
\end{figure*}

In this paper, we propose a novel method, named \textbf{Dream360}, that can generate diverse high-fidelity and high-resolution $360^\circ$ images from a freely given viewpoint (including ceiling, floor, left, right, front and rear view), as shown in Fig.~\ref{fig:coverfig}. Our method is built on the transformer backbone~\cite{Vaswani2017AttentionIA} with two key learning stages: (I) the codebook-based panorama outpainting learns the sphere-specific codebooks from the spherical harmonics (SH) values to represent panoramas with the discrete latent representations and 
to model the distribution of complete panoramas under the condition of given views; (II) the frequency-aware refinement further improves the semantic consistency between given and generated regions, and leverages the frequency-aware consistency loss to restore the structural details and improve the visual fidelity of final outputs (See Fig.~\ref{fig:overview}). 

\begin{figure}[t]
  \centering\includegraphics[width=\linewidth]{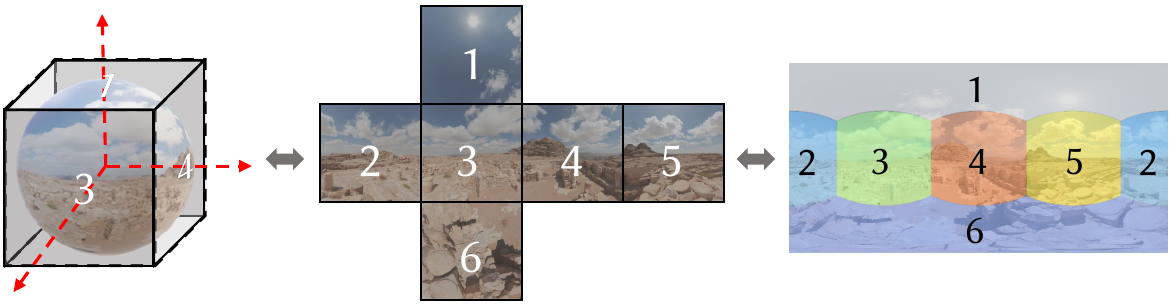}
  \caption{The patches of cubemap projection from a panorama.}
  \label{fig:cubemap}
  \vspace{-10pt}
\end{figure}

Specifically, the codebook learning in the vanilla VQGAN~\cite{Esser2021TamingTF} uniformly down-samples images to obtain discrete latent codes and represent image distribution. However, it is insufficient to directly learn the data distribution of panoramas using VQGAN since the panoramas are spherical and the uniform down-sampling does not maintain the inherent spherical data structure, causing severe artifacts in the reconstruction (See Fig.~\ref{fig:VQGANL}). 
Therefore, in Stage \RN{1}, we first propose Spherical VQGAN (S-VQGAN) that learns the codebook from SH~\cite{fuchs2020se3transformers, Xin2021FastAA}. As SH serves as inherent basis descriptors of the underlying spherical data structure, the learned sphere-specific codebook better represents the discrete data distribution of panoramas for scene modeling using an auto-regressive transformer (See Fig.~\ref{fig:VQGANL}). 
Then, following existing works~\cite{Esser2021TamingTF,Akimoto2022DiverseP3}, we employ an auto-regressive transformer to model the code distribution of complete panoramas under the condition of user-selected views. 
Note following the relations between ERP and cubemap projection(CP) (See Fig.~\ref{fig:cubemap}), we can convert user-selected views into irregular masks of panoramas.

To alleviate the frequency domain gap between the generated region and the given region, we propose a refinement module with a novel frequency-aware consistency loss in Stage \RN{2} to improve the generated results' semantic consistency and visual fidelity. We conduct extensive experiments using outdoor panoramas from the original SUN360~\cite{Xiao2012RecognizingSV} benchmark, where our method significantly outperforms prior-arts. Furthermore, We also conducted a comprehensive user study with 15 participants to evaluate the flexibility and performance of our Dream360. Equipped with VR headsets, these users can freely select a desired direction from the available six options and assess the generated panoramas. The results show that it is challenging to distinguish whether experienced panoramas are generated or real. In summary, our main contributions are \textit{three-fold}:
\begin{itemize}
    \item We propose a flexible and user-centered 360$^\circ$ image outpainting method, which can generate diverse, high-fidelity, and high-resolution panoramas from user-selected viewports.
    \item  We propose S-VQGAN to learn the sphere-specific codebook for panorama outpainting, while we provide a frequency-aware consistency loss to restore more high-frequency details;
    \item Our Dream360 surpasses the existing methods by a significant margin both quantitatively and qualitatively. Users' VR experience in the created virtual scenes further demonstrates the superiority and high interactivity of our Dream360.
\end{itemize} 

\section{Related Work}

\begin{figure*}[ht]
  \centering\includegraphics[width=0.9\linewidth]{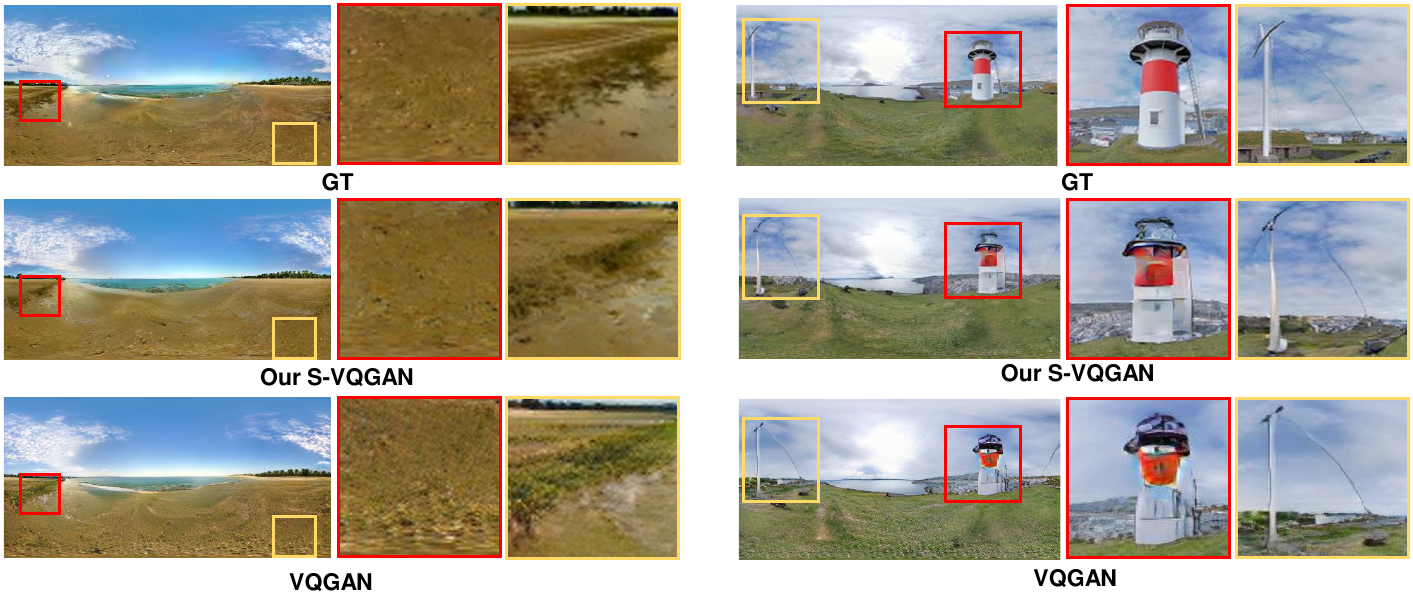}
  \vspace{-5pt}
  \caption{The visual comparison of panorama reconstruction performance at the resolution of $256\times512$ between VQGAN and our S-VQGAN.}
  \label{fig:VQGANL}
  \vspace{-5pt}
\end{figure*}

\noindent\textbf{Planar Image Outpainting.} 
Image outpainting~\cite{Wang2014BiggerPicture,YiWang2019WideContextSI, AlexeiAEfros1999TextureSB, Yang2022SceneGE,Cheng2022InOutDI,Iketani2020AugmentedRI} focuses on creating new contents given a partial region. It is a challenging and ill-posed problem, entailing intensive scene understanding and content consistency maintenance.
Most existing methods~\cite{Cheng2022InOutDI, Pathak2016ContextEF} are based on GANs and follow a typical encoder-decoder framework with adversarial training to enhance photo-realism. However, due to paired training, obtaining diverse results for each masked input is non-trivial. To address this issue, PIC~\cite{Zheng2019PluralisticIC} and UCTGAN~\cite{Zhao2020UCTGANDI} propose dual-path network architectures based on the conditional variational auto-encoders (CVAEs)~\cite{Walker2016AnUF}. Nevertheless, these methods still suffer from low-quality and low-resolution generation. 
Recently, auto-regressive models can naturally support pluralistic outputs; thus, they are used to handle irregular masks and generate diverse results~\cite{Wang2022HighFidelityGI,yu2021diverse}. 
To produce high-resolution results and avoid posterior collapse, VQ-based generative models~\cite{Peng2021GeneratingDS,Esser2021TamingTF,Chang2022MaskGITMG} are often combined with auto-regressive transformers to fill the missing regions.
However, directly applying these methods to our problem ignores the spherical properties of 360 images, and their codebooks are less capable of representing the panoramas (See Fig.~\ref{fig:VQGANL}).

\begin{figure}[t]
  \centering  \includegraphics[width=\linewidth]{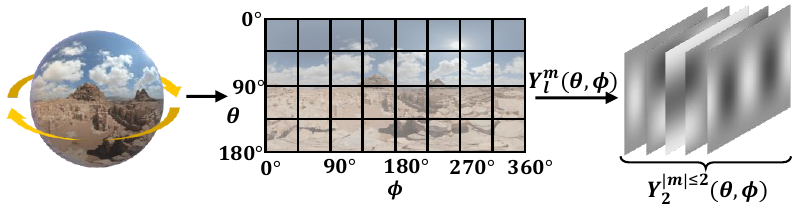}
  \vspace{-15pt}
  \caption{Real-valued spherical harmonics with degree $l=$2 and order $|m|\leq l$ corresponding to a panorama located at the spherical coordinate.}
  \label{fig:SH}
  \vspace{-15pt}
\end{figure}
\noindent\textbf{Panorama Outpainting.} 
It has received increasing attentions owing to its ability to provide immersive experience by allowing users to choose their viewports freely~\cite{ Wang2021BidirectionalSR,Wallgrn2020ACO }. Several works~\cite{Akimoto2019360DegreeIC, Kim2021PaintingOA, Liao2022CylinPaintingS3} leverage the circular property of panoramas to transform the outpainting task into an easier inpainting task and employ GANs to fill the missing pixels. PanoDiff~\cite{Wang2023360DegreePG} first predicts the pose angle of the regional input and then employs the powerful stable diffusion model~\cite{Rombach2021HighResolutionIS} with the text prompt to generate complete panoramas. Recently, IPO-LDM~\cite{Wu2023IPOLDMD3} has also been adopted~\cite{Rombach2021HighResolutionIS} for panorama outpainting and supports multiple types of input masks. However, IPO-LDM heavily relies on expensive depth information to provide structural priors. By contrast, SIG-SS~\cite{Hara2021SphericalIG} proposes a two-stream framework based on CVAE to predict the distribution of partial regions as the condition and jointly predicts the distribution of corresponding complete panoramas. With a set of sparse planar images as input, Sumantri~\etal~\cite{Sumantri2020360PS} proposes a two-stage pipeline that first obtains an incomplete panorama from the input set and then leverages the conditional GAN~\cite{Mirza2014ConditionalGA} with LSGAN~\cite{Mao2016LeastSG} loss to synthesize a complete panorama. PIINet~\cite{Han2020PIINETA3} transforms a masked panorama into multiple cubemap projection (CP) patches and simultaneously predicts the corresponding missing regions on the entire CP patches. However, these methods lack generation diversity and overlook the spherical properties of panoramas. Besides, recent text-driven panorama synthesis method~\cite{Chen2022Text2LightZT} employs the spherical positional embeddings as the conditions of latent code generation and requires text embedding as conditions to synthesize panoramas, which are not applicable to panorama outpainting.

The most relevant work is Omnidreamer~\cite{Akimoto2022DiverseP3}, which employs TT~\cite{Esser2021TamingTF} as the outpainting backbone and an extra AdjustmentNet to up-scale and refine the outpainting results. 
In contrast, our codebook-based outpainting stage is specifically tailored for panoramas. Compared with the poor initialization of codebook learning in~\cite{Akimoto2022DiverseP3}, we investigate the inherent spherical data structure and introduce the concept of sphere-specific codebook learning based on the SH, which offers a superior discrete representation of panoramas. Besides, we explore the frequency gap between the outpainting results and complete target panoramas and propose the frequency-aware consistency loss enables the second stage to synthesize a high-resolution panorama with rich high-frequency information, as depicted in Fig.~\ref{fig:freL}, Fig.~\ref{fig:fremap} and Table~\ref{tab:outpainting}.

\noindent\textbf{Spherical Harmonics (SH).} 
SH refers to an infinite series of complex functions that are single-valued, continuous, and orthonormal basis descriptors on the sphere~\cite{Ramamoorthi2004ASF}. SH has been widely applied to lighting contribution computation~\cite{Ramamoorthi2002AnalyticPC, Xu2020RealtimeIE, Liu2020LightingEV}, scene rendering~\cite{Ramamoorthi2004ASF,Vermast2023Introducing3T, Pessoa2010PhotorealisticRF} and so on. 
Zhu~\etal~\cite{Zhu2020ThePO} employ SH to decompose the panoramas in the frequency domain and extract features in both spatial and frequency domains. In this paper, inspired by the advances in implicit neural representation~\cite{mildenhall2021nerf,Chen2022TransformersAM}, we make the first attempt to learn the codebook from the real SH values to represent the spherical data structure of panoramas. 

\noindent\textbf{Frequency Modeling in Image Generation.} 
A growing body of research explores optimization techniques in the frequency domain to enhance the quality of image generation.
For example, Liu~\etal~\cite{liu2019spectral} use spectral regularization for combating mode collapse in GANs. 
Gal~\etal~\cite{gal2021swagan} design a wavelet-based GAN for image generation.
Jiang~\etal~\cite{Jiang2021FocalFL} introduce the focal frequency loss, which emphasizes hard frequency components. 
Also, several works~\cite{cai2021frequency,jung2021spectral} explore image restoration in the frequency domain.
Inspired by these works, we propose a frequency-aware consistency loss to further improve the visual fidelity of generated results. 

\begin{figure*}[t]
  \centering
  \includegraphics[width=1\linewidth]{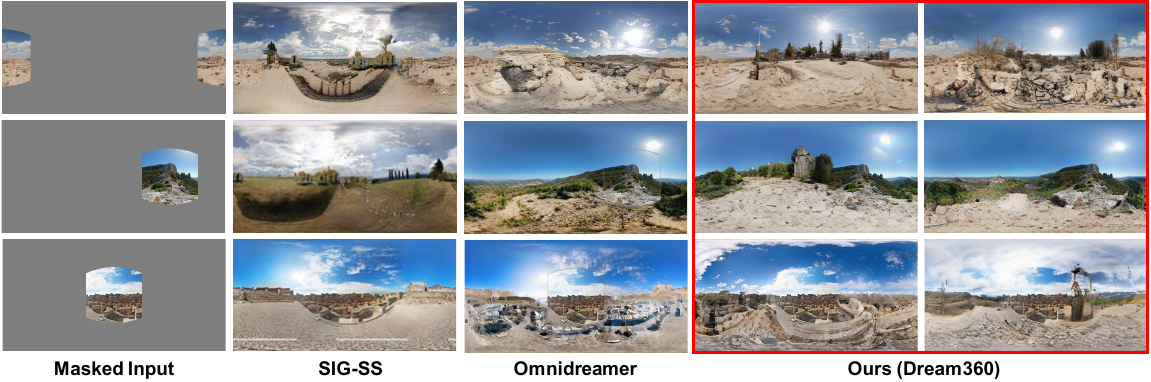}
  \vspace{-15pt}
  \caption{Qualitative results on outdoor scenes of SUN360 dataset. In the last two columns, we show the diverse outpainting results generated by our Dream360. More qualitative examples are shown in Fig.~\ref{fig:resultsL}.}
  \vspace{-15pt}
  \label{fig:outpainting}
\end{figure*}
\section{Preliminary} 
\subsection{Spherical Harmonics for Panoramas}
\label{pre:1}
 As SH provides a comprehensive representation of functions defined on a sphere (see Fig.\ref{fig:SH}), we employ them to learn a sphere-specific codebook for panoramas. Here, we use the real-valued spherical harmonics (RSH) with the degree $l$ and order $m$, as follows:
\begin{equation}
\vspace{-5pt}
    \begin{aligned}
    Y_{l}^{m}(\theta, \phi)=F_{l}^{|m|} * P_{l}^{|m|}&(\cos \theta) *\left\{\begin{array}{ll}
\sin (|m| \phi) & \text { if } m<0 \\
1 / \sqrt{2} & \text { if } m=0 \\
\cos (m \phi) & \text { if } m>0,
\end{array}\right. \\
\end{aligned}
\vspace{-5pt}
\label{eq:sh2}
\end{equation}
where $F_{l}^{m}=\sqrt{\frac{2 l+1}{2 \pi} \frac{(l-m) !}{(l+m) !}}$, $P_{l}^{m}$ is an associated Legendre polynomial~\cite{Xin2021FastAA}, and $(\theta, \phi)$ is the spherical coordinate. Here, $\theta$ denotes the latitude and $\phi$ denotes the longitude.

\subsection{VQGAN}
\label{pre:2}
Vector quantized VAE (VQVAE)~\cite{Oord2017NeuralDR} consists of an encoder-decoder and a codebook $Z \in \mathbb{R}^{K\times d}$ for quantization, where $K$ is the number of embedding vectors (codes), and $d$ is the dimension of codes. Given an image $I \in \mathbb{R}^{H \times W \times 3}$, VQVAE first encodes $I$ into continuous feature maps $z^e \in \mathbb{R}^{h \times w \times d_e}$, $h=H/f, w=W/f$, where $f$ is the scale factor, and $d_e$ is the dimension of feature maps. Subsequently, each pixel feature $ z^e_{(i,j)}, i \in (0,h), j \in (0, w)$  queries its closest codebook entry and obtains its corresponding pixel code $ z^q_{(i,j)}$ as follows:
\begin{equation}
\vspace{-5pt}
     z^q_{(i,j)} = \mathop{\arg\min}\limits_{z^q_{(i,j)} \in Z }\left\|{z}^{e}_{(i,j)}-{z}^q_{(i,j)}\right\|_2^2.
\vspace{-5pt}
\end{equation}
Finally, the quantized code $z^q \in \mathbb{R}^{h \times w \times d} $ is decoded into an image $\hat{I}$ with a reconstruction loss $L_2$  and codebook loss, defined as:
\begin{equation}
         \mathbb{L}_{rec} = \left \| \hat{I} - I \right \|^2, \mathbb{L}_{Codebook} = \left\|\operatorname{sg}\left({z}^{q}\right)-{z}^{e}\right\|_{2}^{2}+\left\|\operatorname{sg}({z}^{e})-{z}^{q}\right\|_{2}^{2},
\label{eq:code}
\end{equation}
where $sg(\cdot)$ denotes the stop-gradient operation.
VQGAN replaces the original $L_2$ loss of VQVAE with a perceptual loss~\cite{Zhang2018TheUE} and introduces adversarial learning, which can enhance the perceptual quality.

\section{Dream360 Framework}
As depicted in Fig.~\ref{fig:overview}, our Dream360 consists of two stages. \textbf{Stage \RN{1}} leverages S-VQGAN to learn two sphere-specific codebooks from the SH values for representing the masked input panoramas and complete target panoramas, respectively. Then, the images are expressed in the sequences of code indices, and a transformer is used to accomplish codebook-based panorama outpainting in an auto-regressive manner. After that, with the up-scaled outpainting results and corresponding high-resolution masked input as the joint inputs, \textbf{Stage \RN{2}} employs a refinement module trained with a novel frequency consistency loss to improve the semantic consistency and visual fidelity while enhancing the resolution of outpainting results. We now describe the details.

\subsection{Codebook-based Panorama Outpainting}
\label{C-HRPO}
\noindent \textbf{S-VQGAN.} As panoramas are directly related to the sphere, the conventional codebook learning with uniform down-sampling ignores the inherent spherical data structure, making it less effective in representing panoramas.
As shown in Fig.~\ref{fig:VQGANL}, the reconstructed panorama of the vanilla VQGAN suffers from severe artifacts and lacks structural details. 
In light of this, our primary task in Stage \RN{1} is to learn a codebook tailored for the panoramas, which can better represent spherical data distribution.
Specifically, we gain prior knowledge from the SH values, the natural basis descriptors defined on a sphere (See Fig.~\ref{fig:SH}), into codebook learning for representing the spherical data structure.
In this way, we address the inadequate codebook learning on panoramas by considering their spherical properties.

As depicted in the first row of Fig.~\ref{fig:overview}, we propose S-VQGAN to learn the codebooks from SH to represent panoramas. Considering the distribution gap between the masked panorama $I^{M}\in \mathbb{R}^{H\times W\times3}$ and complete panorama $I^{C}\in \mathbb{R}^{H\times W\times3}$, we learn two codebooks for the two panoramas, respectively. 
Specifically, taking a masked panorama $I^{M}$,  we first calculate its corresponding SH maps ${SH} \in \mathbb{R}^{ H\times W \times {(D+1)}^2}$ via Eq.~\ref{eq:sh2}, where ${(D+1)}^2$ represents the number of SH at degree $l=D$. 
As $SH$ are only related to the spherical coordinates and independent of pixel values, the $SH$ of $I^{M}$ and $I^{C}$ are identical. 
Then, we employ a mapping network $M_1$ to obtain the codebook $C_{S}^{1}\in \mathbb{R}^{ K \times d}$ from $SH$. The mapping network $M_1$ consists of a down-sample network implemented in the same CNN structure as the encoder of VQGAN and a flatten operator. 
With the learned codebook $C_{S}^{1}$, we follow the process introduced in Sec.~\ref{pre:2} to extract the feature maps $z_e^{M}$ via the encoder $E_1$ and query corresponding codes $z_q^{M}$.
Finally, a decoder $G_1$ is applied to reconstruct the panorama. Similarly, we learn another codebook $C_{S}^{2}$ for the complete panorama $I^{C}$ via a mapping network $M_2$, an encoder $E_2$, and a decoder $G_2$ to reconstruct the complete panorama $I^{C}$.

\noindent \textbf{Diverse panorama outpainting.}
After obtaining two learned codebooks $C_{S}^1$ and $C_{S}^2$, the masked input $I^{M}$ can be represented by a sequence of codebook-indices $s^{M}$ while the complete panorama $I^{C}$ can be represented by a sequence of codebook-indices $s^{C}$. The goal of the diverse panorama outpainting is to predict diverse codebook-indices sequences of the complete panorama based on the codebook-indices sequence of the masked input. Following the previous practices~\cite{Esser2021TamingTF,Akimoto2022DiverseP3}, given sequence $s^{C} \in \mathbb{R}^{L\times1}$ as the conditional input and $s^{M}\in \mathbb{R}^{L\times1}$ as ground truth ($L$ is the sequence length), we train a transformer backbone to learn the distribution of indices mapping from $s^{M}$ to $s^{C}$.
More precisely, the outpainting task can be formulated as given the condition $c$ and indices $s_{<i}$, the transformer will predict the distribution $p(s_i|s_{<i},c)$ of the next index $s_{i}$ and regressively predict the whole sequence $s$. Our Dream360 can easily generate diverse outpainting results by sampling from the learned distribution. 
\textit{Due to the page limit, the inference details can be found in the suppl. material}.

\begin{figure*}[ht]
  \centering
  \includegraphics[width=0.82\linewidth]{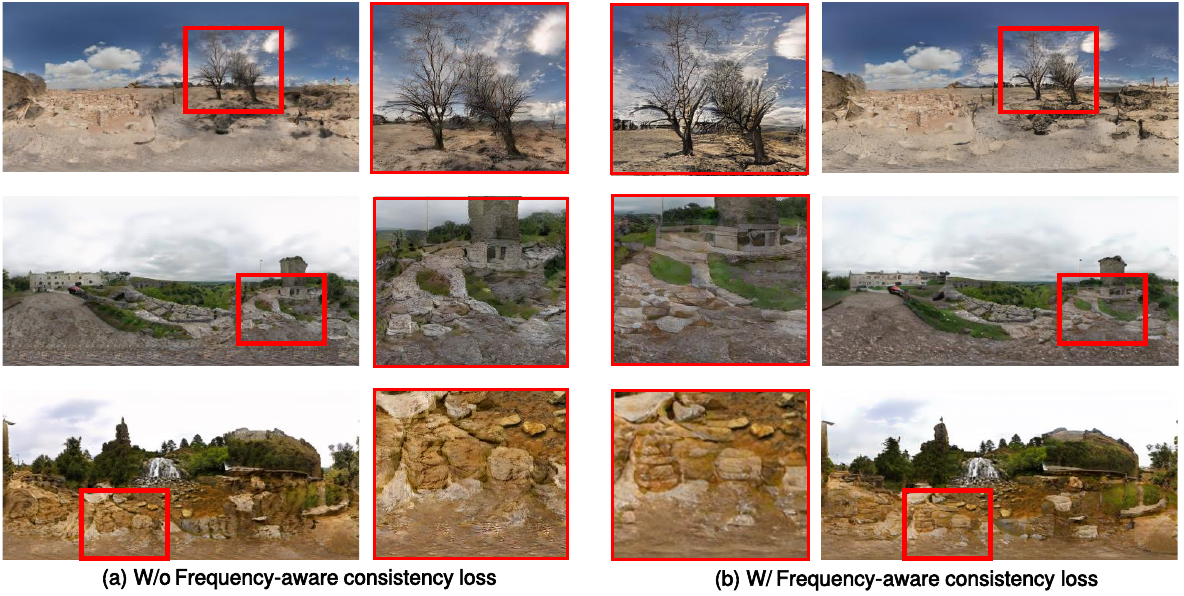}
  \vspace{-5pt}
  \caption{Effect of the frequency-aware consistency loss. (a) represents the results of our S-VQGAN using the refinement module without frequency-aware loss; (b) represents the results with our frequency-aware refinement containing the frequency-aware consistency loss.}
  \vspace{-5pt}
  \label{fig:freL}
\end{figure*}
\begin{figure*}[ht]
  \centering\includegraphics[width=0.9\linewidth]{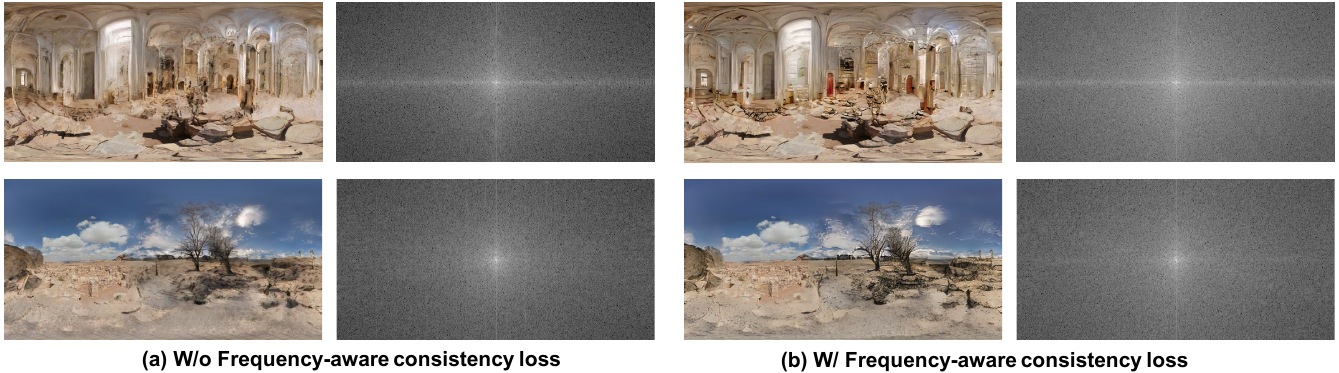}
  \caption{The frequency spectrum with the origin (low frequencies) center shifted and corresponding panoramas in the spatial domain. Our frequency-aware loss can boost the model to learn more high-frequency components (far away from the center).}
  \vspace{-13pt}
  \label{fig:fremap}
\end{figure*}
\begin{figure*}[ht]
  \centering
  \includegraphics[width=0.95\linewidth]{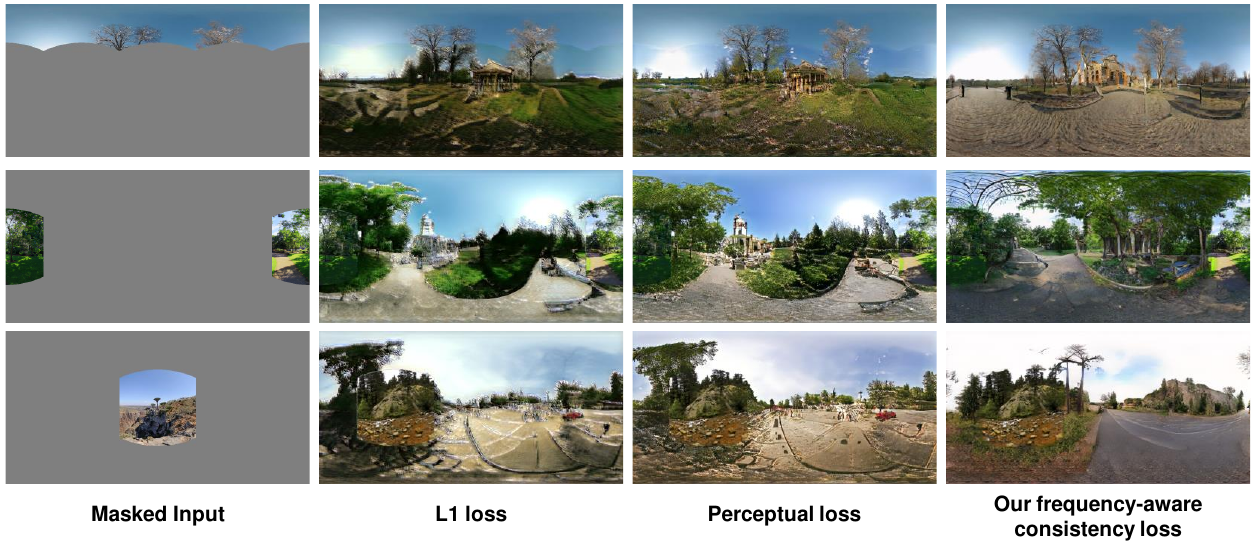}
  \vspace{-10pt}
  \caption{Qualitative comparisons of different loss terms.}
  \vspace{-10pt}
  \label{fig:loss}
\end{figure*}
\begin{figure*}[t]
  \centering
  \includegraphics[width=0.9\linewidth]{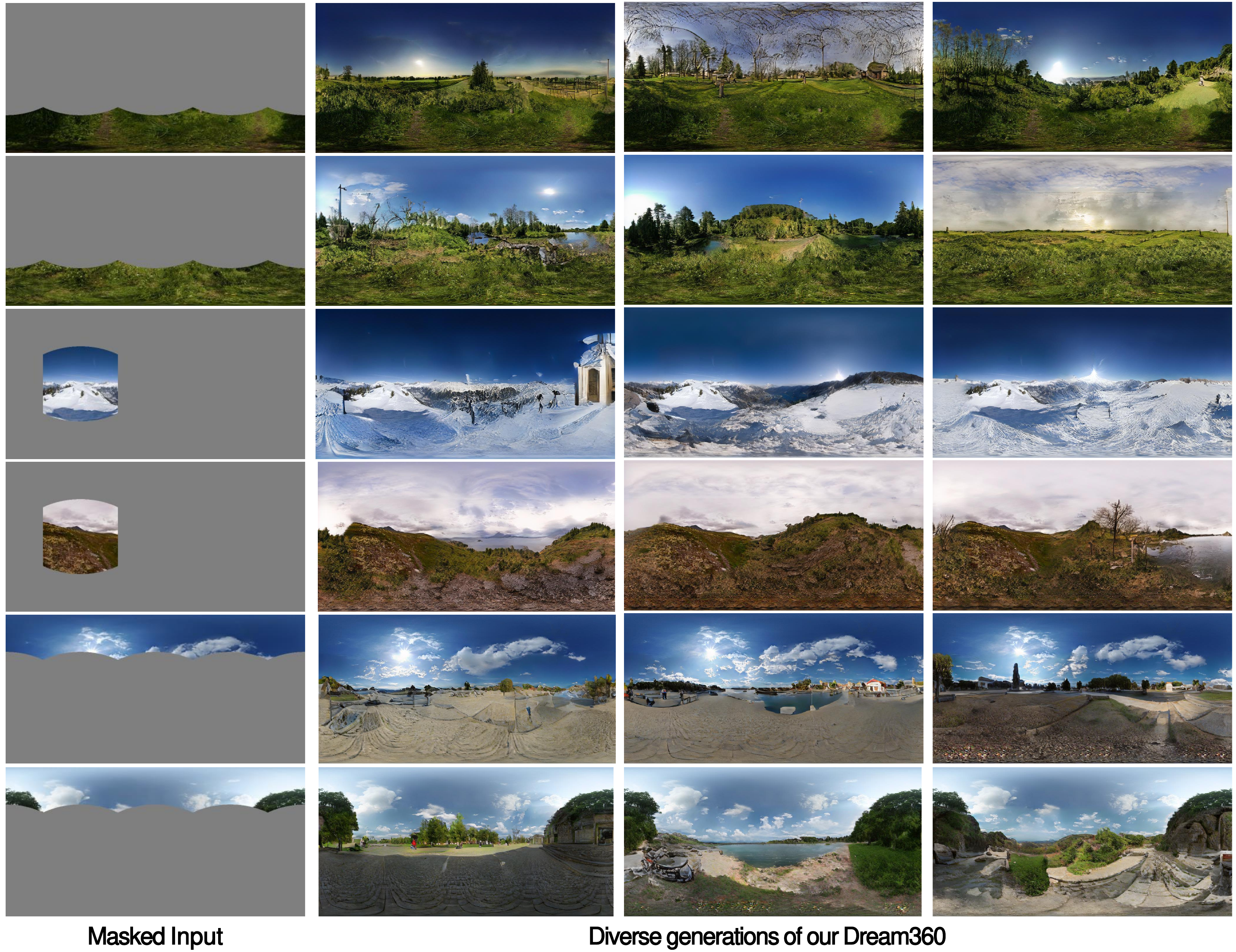}
  \caption{More qualitative results from a narrow field-of-view (NFoV) input with an irregular mask at the ceil, left, and floor location. }
  \label{fig:resultsL}
\end{figure*}

\subsection{Frequency-aware Refinement}
\label{FR}

In Stage \RN{2}, we introduce a refinement module to improve the semantic consistency and visual fidelity of the outpainting results. Specifically, we employ the same U-Net architecture as AdjustmentNet~\cite{Akimoto2022DiverseP3} to refine the up-scaled outpainting results $I_{Up}^{G}$ using high-resolution masked input $I^{M}_{HR}$ as the conditional input. However, when applying the naive AdjustmentNet to our codebook-based panorama outpainting task, the refined results suffer from lacking structural details as demonstrated in Fig.~\ref{fig:freL}(a). 
This deficiency is commonly referred to as spectral bias~\cite{Rahaman2018OnTS,TancikSMFRSRBN20}, which leads generative models to intentionally avoid synthesizing challenging frequency components that are difficult to replicate. 
Therefore, inspired by the existing Fourier analysis~\cite{Jiang2021FocalFL,Fuoli2021FourierSL}, we introduce a frequency-aware consistency loss between the generations and target panoramas into the stage \RN{2} for maintaining holistic semantic consistency and restoring high-frequency component.
Especially our frequency-aware consistency loss employs a GAN training strategy since strictly matching the frequency is insufficient as a single masked input can generate multiple outpainting panoramas (See Fig.~\ref{fig:loss}).

For the frequency-aware consistency loss, we first apply 2D discrete Fourier transform (DFT) to obtain the frequency maps $F^{out}$ and $F^{C}_{HR}$ of the refined panoramas $I^{out}$ and high-resolution target panoramas $I^{C}_{HR}$, formulated as follows:
\begin{equation}
\setlength{\abovedisplayskip}{3pt}
\setlength{\belowdisplayskip}{3pt}
    F(u,v) = \sum_{x=0}^{h-1}\sum_{y=0}^{w-1} p(x,y)\cdot e^{-j2\pi(\frac{ux}{h}+\frac{vy}{w})},
\label{eq4}
\end{equation}
where $(x,y)$ is the pixel coordinate in the spatial domain, $p(x,y)$ is the corresponding pixel value\footnote{
The formulas in this section are applied to gray-scale images and can easily be extended to color images by processing each channel separately.}, $(u,v)$ is the coordinate of a spatial frequency on the frequency spectrum, and $F(u,v)$ is the frequency value. The input image size is $h\times w$, where $h=2H$ and $w=2W$. $e$ and $i$ are Euler's number and the imaginary unit, respectively. Furthermore, each component $F(u,v)$ can be decomposed as two elements: amplitude $\left | F(u,v) \right |$ and phase $\angle F(u,v)$, defined as:
\begin{equation}
\vspace{-3pt}
\begin{aligned}
        \left | F(u,v) \right | &= \sqrt{R(u,v)^2+I(u,v)^2},\\
    \angle F(u,v) &= arctan(\frac{I(u,v)}{R(u,v)}),
\end{aligned}
\vspace{-3pt}
\label{eq5}
\end{equation}
where $R(u,v)$ and $I(u,v)$ are real and imaginary part of $F(u,v)$, respectively.  Then we introduce the discriminator-based frequency-aware consistency loss based on the amplitudes and phases: 
\begin{equation}
        \mathbb{L}_{con}^{F} = \frac{1}{2}\mathbb{L}(|F^{R}|, |F^{C}_{HR}|)+\frac{1}{2}\mathbb{L}(\angle F^{R}, \angle F^{C}_{HR}),
\end{equation}
where $\mathbb{L}(x,y)= log D(y) +log (1 - D(x))$ and $D$ represents the patch-based discriminator~\cite{Isola2016ImagetoImageTW}.
\begin{table}[t] \centering 
  \caption{Reconstruction quality of complete panoramas with different resolution sizes. The best performance is highlighted in red. }
  \label{tab:diff_result}
  \resizebox{0.9\linewidth}{!}{ \begin{tabular}{ccccc}
    \toprule
   Resolution &  Method  &WS-PSNR$\uparrow$&FID$\downarrow$& Codebook Usage$\uparrow$\\
    \midrule
    \multirow{2}{*}{128$\times256$ }&VQGAN &  12.00 & 19.53& 7.8$\%$\\
    & S-VQGAN &  \cellcolor{red!30} 12.57&\cellcolor{red!30} 17.31&\cellcolor{red!30} 10.9$\%$\\
    \midrule
    \multirow{2}{*}{256$\times$512 }&VQGAN &  13.11& 16.22&11.2$\%$\\
    & S-VQGAN&\cellcolor{red!30}13.72&\cellcolor{red!30}12.64&\cellcolor{red!30}19.9$\%$\\
    \midrule
    
    \multirow{2}{*}{512$\times$1024 }& VQGAN &  12.44 & 20.34&13.7$\%$\\
    & S-VQGAN & \cellcolor{red!30}12.97& \cellcolor{red!30}15.58 &\cellcolor{red!30}30.1$\%$\\
  \bottomrule
\end{tabular}}
\vspace{-10pt}
\end{table}
\subsection{Training}
\noindent\textbf{Codebook-based panorama outpainting.} For our S-VQGAN, we train with the same loss functions as the VQGAN, as follows:
\begin{equation}
    \mathbb{L}_{S-VQGAN} = \mathbb{L}_{rec}+ \mathbb{L}_{Codebook}+\lambda\mathbb{L}_{GAN},
\end{equation}
where $\mathbb{L}_{rec}$ is the perceptual loss~\cite{Zhang2018TheUE}, $\mathbb{L}_{codebook}$ is the codebook loss (Eq.~\ref{eq:code}) and $\mathbb{L}_{GAN}$ is the discriminator loss with the patch-based discriminator. The mapping network for codebook learning is optimized by the codebook loss  $\mathbb{L}_{codebook}$. $\lambda$ is the adaptive weight ~\cite{Esser2021TamingTF}. To train the transformer, we directly maximize the conditional log-likelihood $p(s|z_{q}^M)$ as $\mathbb{L}_{trans} =\mathbb{E}_{z^{C}_{q}\sim p(z^{C}_{q})}[-log p(s|z_{q}^M) ].$

\noindent\textbf{Frequency-aware refinement.} The objective function for the refinement stage is defined as follows:
\begin{equation}
    \mathbb{L}_{Refine} = \mathbb{L}_{rec}+ \mathbb{L}_{GAN}+\mathbb{L}_{con}^{F},
    \label{loss:refine}
\end{equation}
where $\mathbb{L}_{rec}$ and $\mathbb{L}_{GAN}$ provide the supervision in the RGB domain, and 
$\mathbb{L}_{con}^{F}$ narrows the gaps in the frequency domain.

\section{Experiments}
\label{experiment}
\subsection{Datasets and Experimental Settings}
\label{datasets}
\noindent\textbf{Datasets and metrics.}
We conduct experiments on outdoor panoramas from the popular dataset, SUN360 dataset~\cite{Xiao2012RecognizingSV}, following~\cite{Jin2020SunskyME}. We numerically evaluate the reconstruction quality using the weighted-to-spherically-uniform PSNR (WS-PSNR) (a specific quality metric for panoramas)~\cite{Sun2017WeightedtoSphericallyUniformQE} and Frechet Inception Distance (FID)~\cite{Heusel2017GANsTB,parmar2021cleanfid} between reconstructed images and original images. Besides, we evaluate the quality of the outpainting generations using FID. \textit{More details can be found in the suppl. material.}

\begin{table}[t] \centering
  \caption{Quantitative outpainting results compared with SoTA methods. * represents Dream360 without the frequency-aware consistency loss.}
  \label{tab:outpainting}
  \resizebox{0.65\linewidth}{!}{ \begin{tabular}{cccc}
    \toprule
    SIG-SS & Omnidreamer & Ours*& Ours \\
    \midrule
   83.13&65.34 &52.67&\cellcolor{red!30}44.89\\
  \bottomrule
\end{tabular}}
\end{table}
\begin{table}[t] \centering
  \caption{Impact of the degree $l$ of SH. The top two performances are highlighted in red and yellow, respectively.}
  \label{tab:degree}
  \resizebox{0.75\linewidth}{!}{ \begin{tabular}{cccc}
    \toprule 
    Method & PSNR $\uparrow$ & WS-PSNR$\uparrow$ &FID$\downarrow$\\
    \midrule
    VQGAN & 18.71& 13.11& 16.22\\
    \midrule
    S-VQGAN ($l$ = 0) & 18.61 & 12.92&15.84\\
    S-VQGAN ($l$ = 1) & 18.87& 13.25&15.74\\
    S-VQGAN ($l$ = 3) & \cellcolor{red!30}19.42 &\cellcolor{red!30}13.72&\cellcolor{red!30}12.64\\
    S-VQGAN ($l$ = 5) & \cellcolor{yellow!30}19.25& \cellcolor{yellow!30}13.55&\cellcolor{yellow!30}14.70\\
    S-VQGAN ($l$ = 7) & 18.46&12.88&15.61\\
  \bottomrule
\end{tabular}}
\vspace{-10pt}
\end{table}

\noindent\textbf{Implementation details.} In this work, we achieve diverse outpainting results from an NFoV input with \textbf{an irregular mask at one position of the available six options}, which is more challenging than prior arts that mainly focus on the centrally located input with a rectangular mask. Note, in this work, we follow the previous methods~\cite{Hara2021SphericalIG,Akimoto2022DiverseP3} and ignore the case when a masked input has a tilting camera orientation. More specifically, we apply the cubemap projection (CP) to obtain the flexible masks on the ERP-type panoramas (See Fig.~\ref{fig:cubemap}). We train Dream360 stage by stage on a single NVIDIA V100 GPU and use the Adam optimizer~\cite{Kingma2014AdamAM} with a learning rate of $4.5e^{-6}$ and batch size of $4$ for all networks. We train the transformer of Stage \RN{1} with 20 epochs and the remaining networks with 30 epochs. We use masked panoramas of $256\times512$ as the inputs and finally produce high-resolution panoramas of $512\times1024$. \textit{Due to the page limit, more implementation details can be found in the suppl. material.}

\subsection{Evaluations on Diverse High-fidelity Outpainting}
We further evaluate the capability of our Dream360 in directly generating diverse high-resolution and high-fidelity panoramas from a single NFoV input with an irregular mask at an arbitrary location. We calculate the average FID score of whole patches as the final score. In Table~\ref{tab:outpainting}, we present the quantitative results compared with existing SoTA methods. Especially, as our Dream360 is built upon the transformer-based backbone, we mainly consider the transformer-based baselines for comparisons. For the outdoor scenes of the SUN360 dataset, Our Dream360 outperforms SIG-SS by 46$\%$ and outperforms Omnidreamer by 31.29$\%$. Besides, the visual comparisons are shown in Fig.~\ref{fig:outpainting}. Our Dream360 can generate diverse high-fidelity panoramas with better holistic semantic consistency, and alleviate the sharp boundary between the input regions and generated regions. Moreover, our proposed frequency-aware consistency loss can boost the model to learn more high-frequency components (See Fig.~\ref{fig:fremap}) and restore more local structural details (See Fig.~\ref{fig:freL}),~\eg, buildings, trees, and stones.
\begin{table}[t] \centering
  \caption{The ablation results for different losses based on the frequency domain. Our loss means our frequency-aware consistency loss based on the GAN training strategy.}
  \label{tab:freq}
  \resizebox{0.7\linewidth}{!}{\begin{tabular}{cccc}
    \toprule
     & L1 loss& Perceptual loss &Our loss\\
    \midrule
    FID$\downarrow$  &57.22 &52.27 & \cellcolor{red!30}44.89\\
  \bottomrule
\end{tabular}}
\vspace{-10pt}
\end{table}

\subsection{Ablation Study}
\noindent\textbf{Degree of SH.} We now compare the panorama reconstruction performance with various degrees of SH on the SUN360 test dataset. As observed from Table~\ref{tab:degree}, starting from $l$=0, the reconstruction quality first improves with the increase of $l$ and then drops significantly. The result indicates that too large a degree leads to difficulty in learning the representative codebook. For this reason, we choose $l=3$ as the default degree of SH for other experiments.

\noindent\textbf{Frequency-aware consistency loss.}
As introduced in Sec.~\ref{FR}, we aim to narrow the difference in frequency space between the real panoramas and refined generations. Several other loss functions exist to supervise frequency information,\ie, L1 loss and perceptual loss. In Table~\ref{tab:freq} and Fig.~\ref{fig:loss}, we compare our consistency loss with the L1 loss, perceptual loss based on the frequency domain on the SUN360 test dataset. Overall, our frequency-aware consistency loss based on the discriminator loss achieves the best performance.
\begin{table}[t] \centering
  \caption{The ablation results for different locations.}
  \label{tab:location}
  \resizebox{0.95\linewidth}{!}{ \begin{tabular}{ccccccc}
    \toprule
     & 1  &2&3 &4&5&6\\
    \midrule
    Omnidreamer & 72.82& 60.56& 63.61& 64.69&64.80 & 74.76\\
    Dream360&47.19 & 39.85& 41.23&46.75 &45.17 &48.96\\
  \bottomrule
\end{tabular}}
 \vspace{-15pt}
\end{table}

\noindent\textbf{Locations of input region.} As shown in Fig.~\ref{fig:coverfig}, taking the masked panorama projected by the cubemap patches as the input, we can obtain the masks at arbitrarily one of the 6 locations. The difficulty of panorama outpainting depends on the locations of the masks. In Table~\ref{tab:location}, we compare our performance with Omnidreamer at each input location, and Fig.~\ref{fig:resultsL} validates that our Dream360 can handle various conditions.

\subsection{Extensions}

\begin{figure}[t]
  \centering
\includegraphics[width=0.9\linewidth]{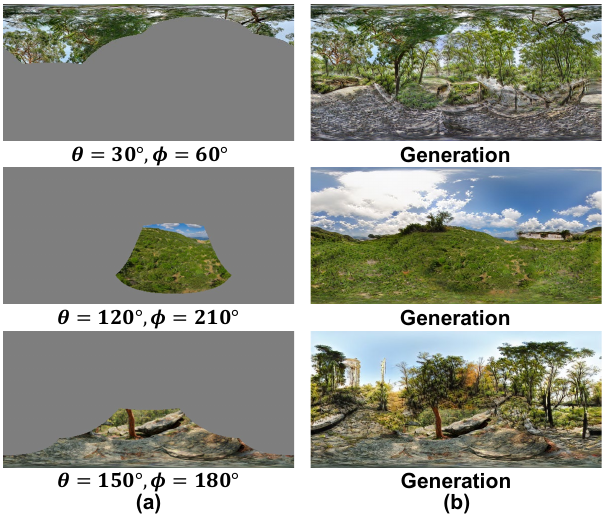}
\vspace{-5pt}
  \caption{(a) The masked panoramas corresponding to the more flexible viewpoints, provided by tangent projection (Especially, we set the filed-of-view (FoV) as 120$^\circ$ $\times$120$^\circ$); (b) Generations of our Dream360 from the masked panoramas in (a).}
  \label{fig:exten_tp}
  \vspace{-10pt}
\end{figure}

\begin{figure}[t]
  \centering
\includegraphics[width=1\linewidth]{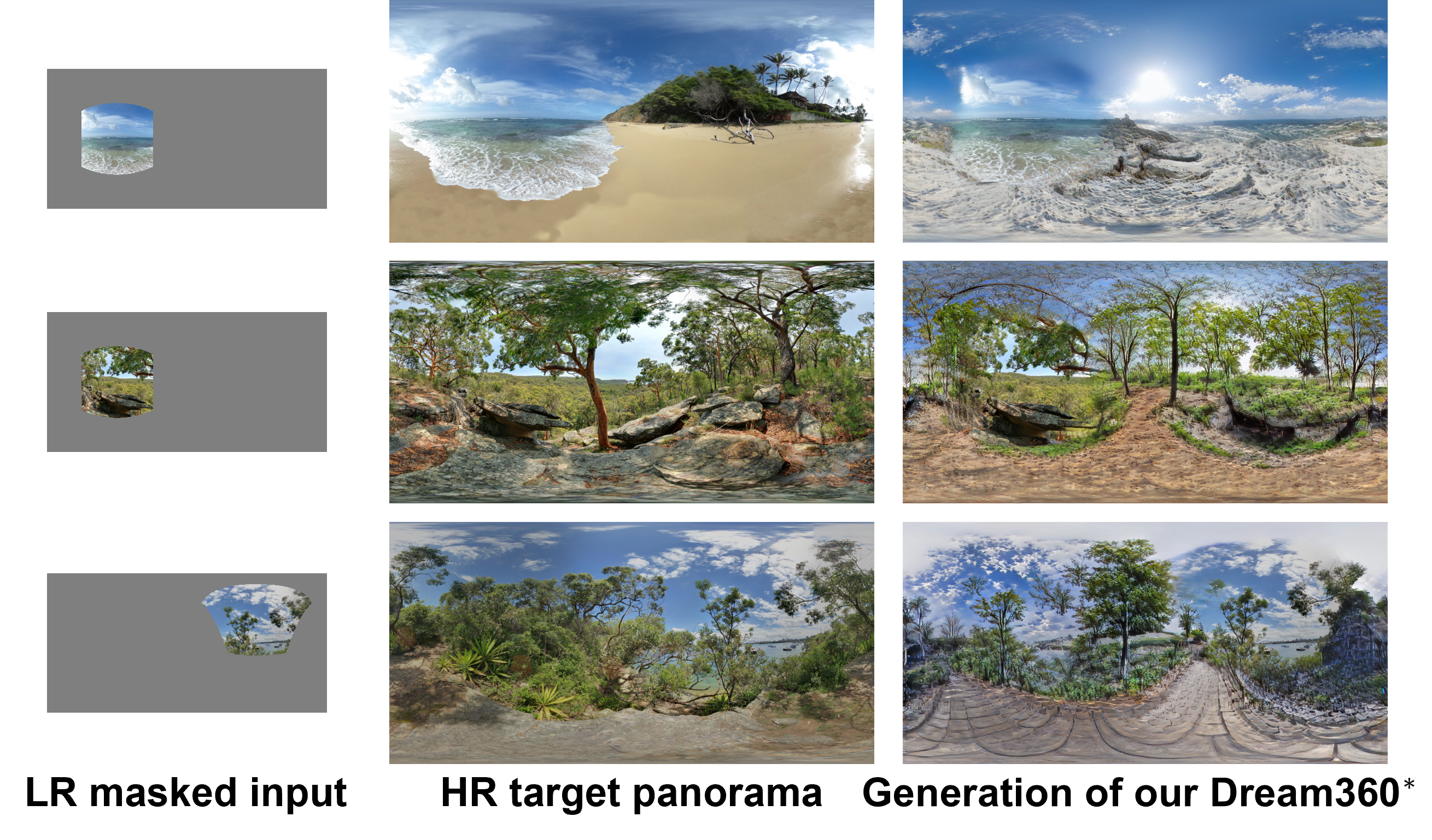}
\vspace{-15pt}
  \caption{Qualitative results with the higher resolutions. The masked inputs are with the resolution size of $256\times512$, while the generation outputs of our Dream360 are with $1024 \times 2048$. $*$ means that the stage \RN{1} is trained on SUN360, and the stage \RN{2} is trained on Flickr360 dataset.}
  \label{fig:exten_hr}
  \vspace{-10pt}
\end{figure}

\noindent\textbf{Extension with more flexible viewpoints.} Our Dream360 exhibits superior versatility, capable of handling more flexible viewpoint directions. Specifically, we introduce tangent projections (TP)~\cite{Eder2019TangentIF,Ai2023HRDFuseM3} to offer masks corresponding to various flexible viewing directions. As depicted in Fig.~\ref{fig:exten_tp}(a), by altering the centers and field-of-view (FoV) sizes of TP patches, we obtain the diverse masks on the ERP panoramas. Moreover, as demonstrated in Fig.~\ref{fig:exten_tp}(b), after re-training Dream360 using these masked panoramas, we generate diverse, high-fidelity complete panoramas.

\noindent\textbf{Extension with higher generation resolutions.} As introduced in Sec.~\ref{FR}, our frequency-aware refinement stage refines the up-scaled outpainting results $I_{Up}^{G}$ with the high-resolution masked input $I_{HR}^{M}$. Considering the resolution sizes of panoramas in the SUN360 dataset is at $512\times1024$, \ie, $I_{HR}^{M} \in \mathbb{R}^{512\times1024\times3}$, our Dream360 takes masked images with the resolution size of $256\times 512$ as input to generate complete panoramic images at $512\times1024$, which is consistent with the existing SoTA method~\cite{Akimoto2022DiverseP3}. Notably, while providing higher-resolution target ground truths, we only need to retrain the stage \RN{2} to achieve higher-resolution generations. As illustrated in Fig.~\ref{fig:exten_hr}, leveraging the stage \RN{1} trained on the SUN360 dataset, we continue to train the second stage on the Flickr360 dataset~\cite{Cao2023NTIRE2C}, where panoramas have the resolution size of $1024\times2048$. Observed from the results, our Dream360 easily generates higher-resolution and high-fidelity panoramas.

\begin{figure}[t]
  \centering
\includegraphics[width=0.85\linewidth]{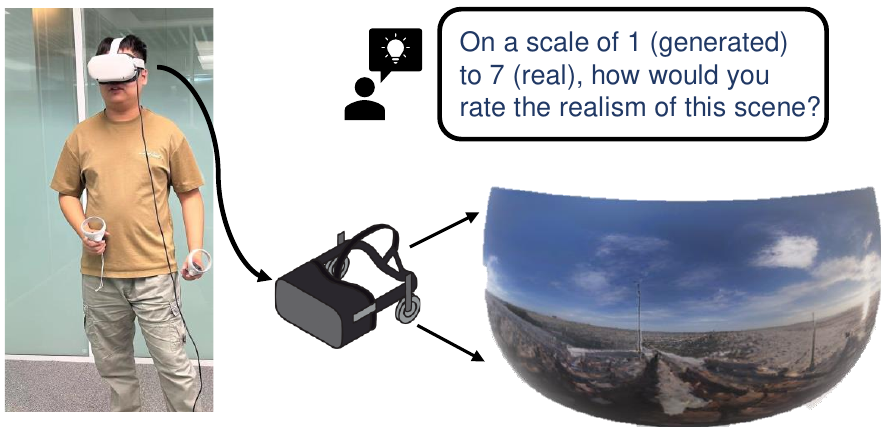}
  \caption{The participants were equipped with Oculus Quest2 and enjoyed virtual tourism using CenarioVR software.}
  \label{fig:vr}
  \vspace{-10pt}
\end{figure}
\begin{figure}[t]
  \centering
  \includegraphics[width=1\linewidth]{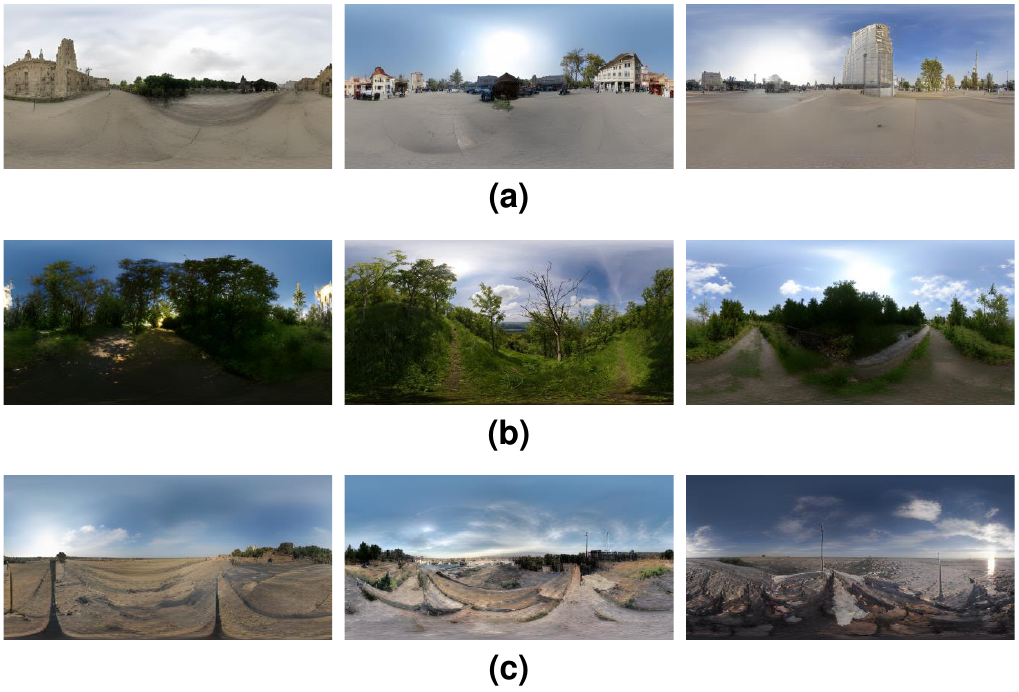}
  \vspace{-15pt}
  \caption{Fake samples for Task 1: (a) Set 1, (b) Set 2, (c) Set 3.}
  \vspace{-15pt}
  \label{fig:task1_sample}
\end{figure}
\section{User study}
In order to comprehensively evaluate the quality of the outdoor scenes generated by our method and assess their impact on users' interactive experience, we conducted a user study in a virtual reality environment involving 15 participants. The study consisted of three tasks: a Real/Fake task, a Find Real Part task, and an Enjoy the Watching task. Details of the task design will be elaborated upon in the subsequent sections. Data collection involved a blend of quantitative analysis, incorporating subjective ratings, and qualitative research conducted via semi-structured interviews. 

\subsection{Participants and devices}
We recruited 15 university students (10 males and 5 females) as participants. Notably, 73.33$\%$ of the participants fell within the 18-24 age group, while the remaining 26.67$\%$ belonged to the 25-34 age bracket. Furthermore, 5 participants had no VR experience in the last six months, while 6 participants were not familiar with panoramas. The experiments were conducted using the Meta Quest 2 VR Headset, and the Oculus software CenarioVR was utilized as our player (See Fig.~\ref{fig:vr}). In addition, the participants used two controllers for switching panoramas and selecting viewpoint directions.

\subsection{Task design}
We designed three tasks from both objective evaluation and subjective perception aspects, namely the Real/Fake task, Find Real Part task, and Enjoying the Watching task. In the first two tasks, the goal is to assess the realism of generated parts, while the last task aims to make users focus on the interactivity and satisfaction brought by our Dream360.

\begin{itemize}
    \item \textbf{Real/Fake task.} In this task, we collected a set of outdoor virtual scenes for participants to observe in a VR environment. The set comprised 1 real-world panorama (i.e., "real" sample) and 4 generated panoramas from different directions (i.e., "fake" samples). For each participant, we totally provided three sample sets, and the order of samples was randomized within each set. Participants were instructed to assign a realism score to each scene using a 7-point Likert scale.
    \item \textbf{Find Real Part task.} To further evaluate the fidelity of our generation, we introduced this Find Real Part task. In this task, we collected three sample sets, with each set comprising six distinct scenes generated from different directions, \ie, celling, left, front, right, back, floor. Participants are informed that there exists only one real region in each scenes. We randomized the order of scenes within each set and allowed participants to freely observe them. After watching each scene, participants were asked to identify the real part from the six options. The more challenging it is for participants to discern which part is real, the higher the fidelity of our generated content.
    \item \textbf{Enjoying the Watching task.} In this task, we provide virtual tourism for each participant. Firstly, participants view a real-world scene and then they are asked to select one viewpoint direction as the fixed region. Consequently, in the VR environment, the participants can freely observe the generated virtual scenes according to their chosen viewpoints and compare the generate scenes with the real-world scenes. After observation, they are asked to provide assessments on various aspects.
\end{itemize}
\vspace{-10pt}
\subsection{Measurements} 
In the first task, we assessed the realism of fake samples using a 7-point Likert scale. Additionally, we employed a threshold of 5, where ratings exceeding 5 indicated that participants considered the scene as real, while ratings below 5 are classified as fake. Subsequently, a confusion matrix was computed based on classification accuracy to better illustrate the realism of the scenes generated by Dream360. In the second task, akin to a multi-class classification, we computed a confusion matrix to evaluate participants' performance. For the last task, we utilized a 7-point Likert scale to measure participants' perceptions across several dimensions, including scene realism, visual aesthetics, operability of direction selection, alignment between generated and selected directions, and level of realism in the fusion region. Finally, participants were encouraged to provide feedback regarding their emotional responses and immersive experience during the overall experience.

\subsection{Results}
We present an analysis of participants' scores, impressions and perspectives concerning their VR experience with our proposed Dream360. The following discussion provides an examination of users' feedback across the three tasks.

\begin{figure}[!t]
  \centering
\includegraphics[width=0.95\linewidth]{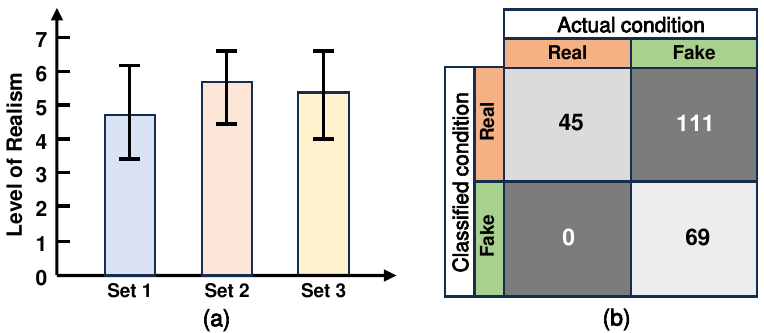}
\vspace{-5pt}
  \caption{The responses from participants in the Real/Fake task (a) Bar charts of the participants’ scores for the realism level of fake samples with different sample sets; (b) The classification confusion matrix of participants' performance (we have totally $15\times4\times3 = 180$ fake samples and $15\times1\times3 = 45$ real samples).}
  \vspace{-10pt}
  \label{fig:t1_results}
\end{figure}

\noindent \textbf{Real/Fake.} Fig.~\ref{fig:t1_results} (a) displays participants' ratings of the realism of virtual scenes, focusing solely on the generated scenes. It is evident that participants consistently rated realism highly, with mean scores surpassing 4.5 for all three sets. Specifically, Set 1 received a mean rating of 4.825 with a standard deviation of 1.412, Set 2 received a mean rating of 5.525 with a standard deviation of 1.109, and Set 3 received a mean rating of 5.375 with a standard deviation of 1.353. Notably, participants assigned lower scores with larger variances to Set 1, which contains more buildings (see Fig.~\ref{fig:task1_sample}). This observation suggests that buildings, with their complicated details, pose a greater challenge for our Dream360 in terms of accurate generation. In contrast, natural landscapes, as shown in Set 2 and Set 3, are more faithfully generated. Furthermore, as shown in Fig.~\ref{fig:t1_results} (b), we computed the confusion matrix to evaluate participants' classification performance using the previously mentioned strategy. Remarkably, over 60$\%$ of the fake samples (total $15\times4\times3 = 180$ fake samples) were classified as real samples, demonstrating the impressive capabilities of our Dream360.
\begin{figure}[t]
  \centering
  \includegraphics[width=0.85\linewidth]{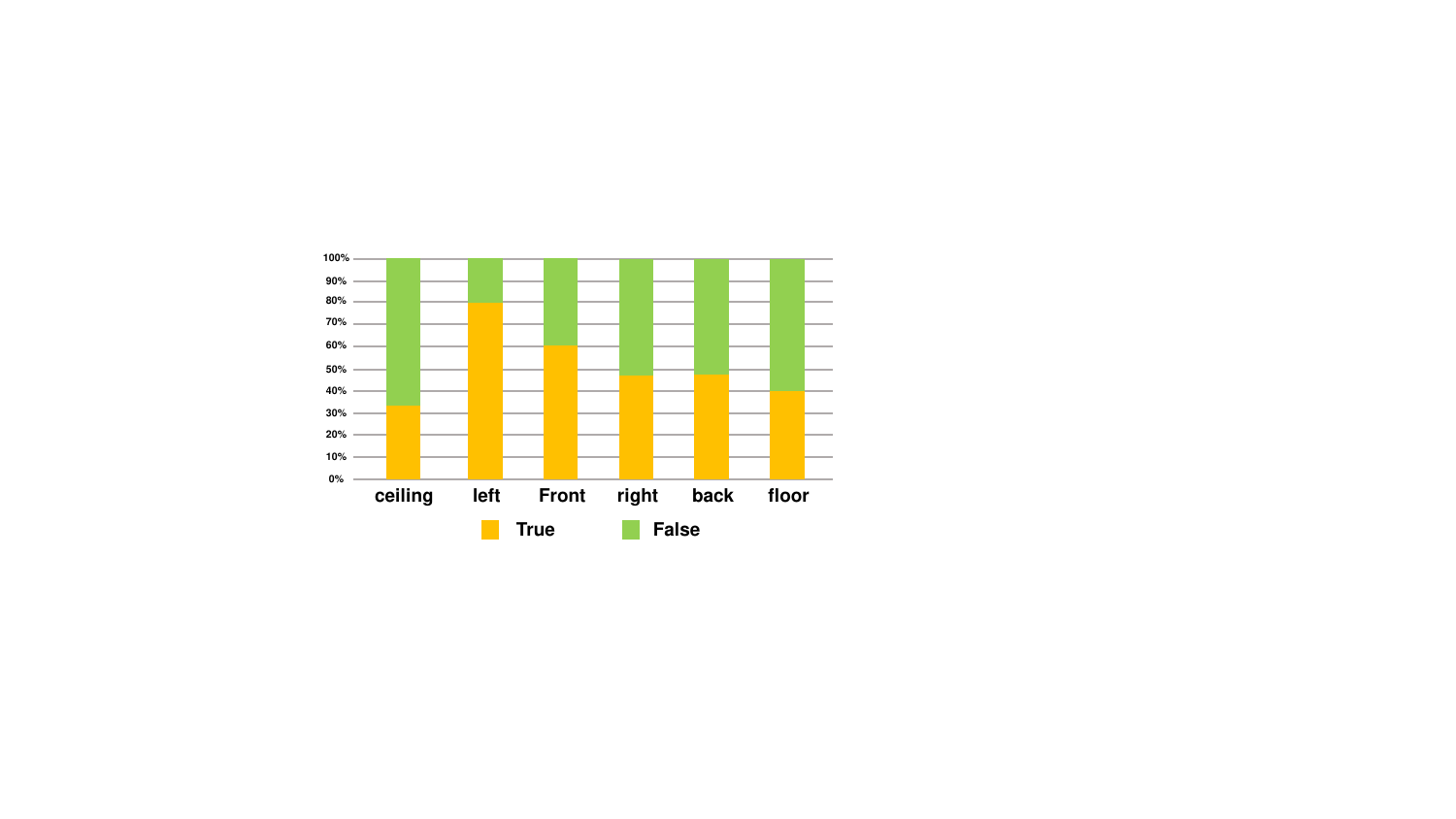}
  \vspace{-8pt}
  \caption{The participants' detection accuracy in the Find Real Part task.}
  \label{fig:t2_results}
    \vspace{-15pt}
\end{figure}

\noindent \textbf{Find Real Part.} As shown in Fig.~\ref{fig:t2_results}, we collected a total of 90 detection results (15$\times$6). Based on the participants' responses, the detection accuracy for each viewpoint direction was calculated, following the relationships depicted in Fig.~\ref{fig:cubemap}. The results indicate that participants achieved the lowest detection accuracy for the ceiling and floor viewpoint directions. This observation can be attributed to participants' observational biases, as they tend to overlook contents on the ceiling and floor. Participants exhibited the highest detection accuracy for the left viewpoint direction, which is located across the left-right boundary and is easy to recognize. For the other three directions, the average detection accuracy is approximately 55$\%$, highlighting an element of deception in our generated content.
\begin{figure}[t]
  \centering
  \includegraphics[width=0.8\linewidth]{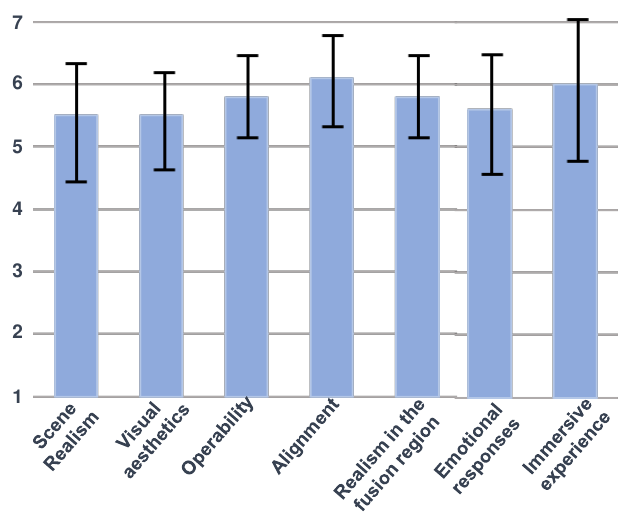}
   \vspace{-10pt}
  \caption{The responses from participants on various aspects of the VR experience using a 7-point Likert scale.}
  \label{fig:t3_results}
      \vspace{-15pt}
\end{figure}

\noindent \textbf{Enjoy the Watching.} In Fig.~\ref{fig:t3_results}, we conducted an analysis of participants' evaluations across seven key aspects: scene realism, visual aesthetics, operability of viewpoint direction selection, alignment between the generated content and selected viewpoints, the level of realism in the fusion region, emotional responses, and the overall immersive experience during watching. The results revealed that participants consistently provided scores above 5, indicating a positive assessment. Notably, the alignment between the generated content and the specified viewpoint directions received an average rating of 6.1, with a standard deviation of 0.782. This suggests that our Dream360 adeptly perceives the provided content and offers coherent completion. Furthermore, participants highly rated the immersive experience while viewing (average score of 6) and the interactivity (average score of 5.8), underscoring Dream360's capacity to enhance the virtual experience for users in VR applications.

\section{Limitations and Future work}
Transformer-based Dream360 fails to adapt to inputs of any scale with one-time training. Applying Dream360 to data with different resolutions requires retraining, which is complex and time-consuming. Therefore, in the future, we will consider exploring a diffusion-based adaptive panorama outpainting method that can be tailored to various resolutions [T-K: cite IDM(Implicit Diffusion Models)? typically diffusion methods work only for fixed resolutions. Oh you mention this below]. Meanwhile, we will also explore how to utilize existing extensive datasets of planar images for panorama outpainting. We have proposed frequency-aware refinement to enhance the high-frequency details but it is also limited to intricate textures, especially buildings. Therefore we will attempt to explore the integration of more fine-grained generators as a potential solution to this problem in the future work. Furthermore, participants' feedback underscores the significance of panorama resolution to the overall experience. Although we could generate outputs at a resolution of 512$\times$1024 or higher, by providing the high-resolution target panoramas, the existing high-resolution panorama datasets are limited and insufficient for VR applications. To address this and consider computational constraints, we may explore the combination of implicit neural representation-based super-resolution methods and diffusion-based panorama outpainting methods in the future work. This approach endeavors to achieve 4K-level panorama generation.

\section{Conclusions}
This paper proposed a novel solution for diverse high-fidelity and high-resolution panorama outpainting, given an irregularly masked input at an arbitrary location. To address the two issues: 1) lack of full consideration of spherical properties of a 360$^\circ$ image; 2) limitation of improving the holistic semantic consistency and restoring the local structural details, our Dream360 introduced S-VQGAN to learn the sphere-specific codebook from SH, employed the codebook-based outpainting process to predict diverse outpainting results and provided a frequency-aware refinement to enhance visual fidelity of the generated results. Quantitative and qualitative results showed that the codebook learning from SH increases the codebook usage and improves the reconstruction quality, and the frequency-aware refinement, which derives the model focusing on the frequency components, significantly enhances the visual fidelity. Especially, we conducted a series of comprehensive user studies to evaluate Dream360's performance in VR applications. The results clearly demonstrated that Dream360 allows users to get a more interesting and immersive VR experience.


\bibliographystyle{abbrv-doi-hyperref}

\bibliography{template}

\appendix 

\section{About Appendices}
Refer to \cref{sec:appendices_inst} for instructions regarding appendices.

\section{Troubleshooting}
\label{appendix:troubleshooting}

\end{document}


\firstsection{Datasets and Metrics}
\maketitle
\subsection{Datasets}
Note that the original SUN360 dataset, which contains 65,000 panoramas, is not fully available during the time of this paper's writing. Therefore, we utilized a subset of SUN360 following~\cite{Jin2020SunskyME}, comprising a total of 22,126 panoramas from both the ``Outdoor'' classes (refer to Fig.~\ref{fig:sup_dataset} (a)). This subset is further divided into 20,000 training images and 2,126 test images.

\subsection{Metrics} We employ the
weighted-to-spherically-uniform PSNR (WS-PSNR) \cite{Sun2017WeightedtoSphericallyUniformQE} and Frechet Inception Distance (FID)~\cite{Heusel2017GANsTB,parmar2021cleanfid} to measure the reconstruction quality and outpainting quality. Specifically, we use the clean-FID as the script to calculate the FID. 

\begin{figure}[h]
  \centering
  \includegraphics[width=0.8\linewidth]
  {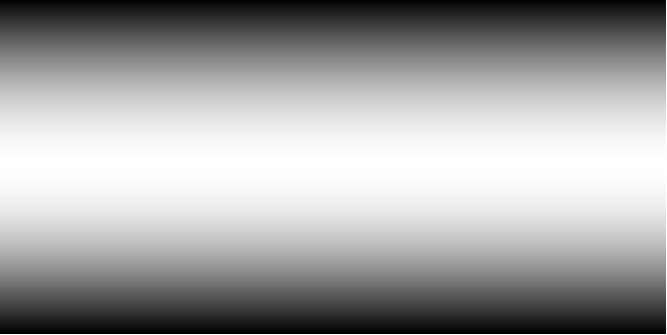}
    \vspace{-10pt}
  \caption{The visualization of weight map in the WS-PSNR.}
  \vspace{-10pt}
  \label{fig:supp_weight}
\end{figure}

For WS-PSNR, it uses the scaling factor $w(i, j)$ from the 2D plane to the sphere as a weighting factor for PSNR computation, as follows:
\begin{equation}
    w(i,j) = cos\frac{(j+0.5-N/2)\pi}{N}
\end{equation}
where $i\in (1,M)$, $j\in (1,N)$ and $M=2N$. As shown in Fig.~\ref{fig:supp_weight}, the weight map $w$ can be viewed as an attention map that mainly focuses on the equator.
FID calculates the feature distribution distance between completion results and natural images and is widely used as an evaluation metric for reconstruction and outpainting tasks.

\section{Implementation Details}
\noindent \textbf{Diverse panorama outpainting inference.} 
During the inference in \textbf{Stage \RN{1}}, we use circular inference~\cite{Akimoto2022DiverseP3} to obtain completed 360-degree images. After predicting the sequence of outpainting results, we first map indices of a sequence back to their corresponding codebook entries and reshape the codes to obtain the quantized feature map with the original resolution size. 
Then we duplicate both ends of the quantized feature map on the opposite side prior to the transformer estimation process. This duplication allows for a seamless connection between the two ends of the feature map in higher-order features.
During the estimation phase, the transformer operates in raster order, sequentially estimating each row. 
After estimating a row, the estimation results from both ends of the quantized feature map are copied from the opposite side and substituted into their respective positions. 
This circular estimation approach helps maintain connectivity between the two ends of the feature map throughout the transformation process.
By incorporating circular estimation with a transformer, we achieve improved connection at the semantic level when decoding the feature map into an image. 
This enhanced connectivity leads to a more plausible completion of the 360-degree image, as it ensures smooth transitions and consistency across the entire image.
Thus, the overall inference is 
\begin{equation}
    y = G_{\text{Adjust}}(G_{2}(T(E_{1}(x))), x),
\end{equation}
where $G_{\text{Adjust}}$, $G_{A}$, $T$, and $E_{B}$ indicate AdjustmentNet, the decoder of VQGAN trained on completed panorama image, the transformer, and the encoder of VQGAN trained on masked panorama image, respectively.

\noindent \textbf{Frequency-aware refinement stage.}
The network architecture of our refinement module is a U-Net structure implemented in the same CNN structure as VQGAN without the VQ mechanism. We employ the same training strategy with the AdjustmentNet of Omnidreamer~\cite{Akimoto2022DiverseP3}. We train the refinement module by restoring the combinations of preprocessed images and masked inputs to the complete targets and inference directly using outpainting results and high-resolution masked inputs. Specifically, the preprocessing includes three steps: 1) reconstructing the complete targets via learned S-VQGAN and these reconstructed images are different from the targets, which can view the ``alternatives'' of outpainting results; 2) adding color jitter to the complete targets before reconstruction, which results in the color difference between the targets and reconstructed images; 3) scaling down the complete targets before the reconstruction and up-scale the reconstructed images to the original scale. Note that, we use the combination of preprocessed images and masked targets as the joint input and train the U-Net to restore the complete targets with the objective function (Eq.~9 in the main paper). For instance, given the complete target $I^{C}\in \mathcal{R}^{256\times512\times3}$, we first down-scale it to $128\times256\times3$ and add color jitter to it, and then we use the learned S-VQGAN to obtain the reconstruction $\hat{I^{c}}\in \mathcal{R}^{128\times256\times3}$. After up-scaling $\hat{I^{C}}$ to $256\times512\times3$, we combine it with masked input $I^{M}$ and use the U-Net to predict the refined output $I^{out} \in 256\times512\times3$ supervised by complete target $I^{C}$. During inference, we can predict the outpainting results $I^{G}\in \mathcal{R}^{256\times512\times3}$ and up-scale it to obtain $I_{Up}^{G} \in \mathcal{R}^{512\times1024\times3}$. By adding high-resolution masked input $I_{M}^{HR} \in \mathcal{R}^{512\times1024\times3}$, we leverage the U-Net to output high-fidelity and high-resolution final output.

\section{Additional Experimental Results}
In this section, we will show more
experimental results for panorama reconstruction, diverse high-fidelity, and high-resolution panorama outpainting, qualitative comparisons to existing SoTA methods, and ablation studies for our key design. 
\subsection{Panorama Reconstruction}
\begin{figure*}[ht]
  \centering
  \includegraphics[width=0.7\linewidth]
  {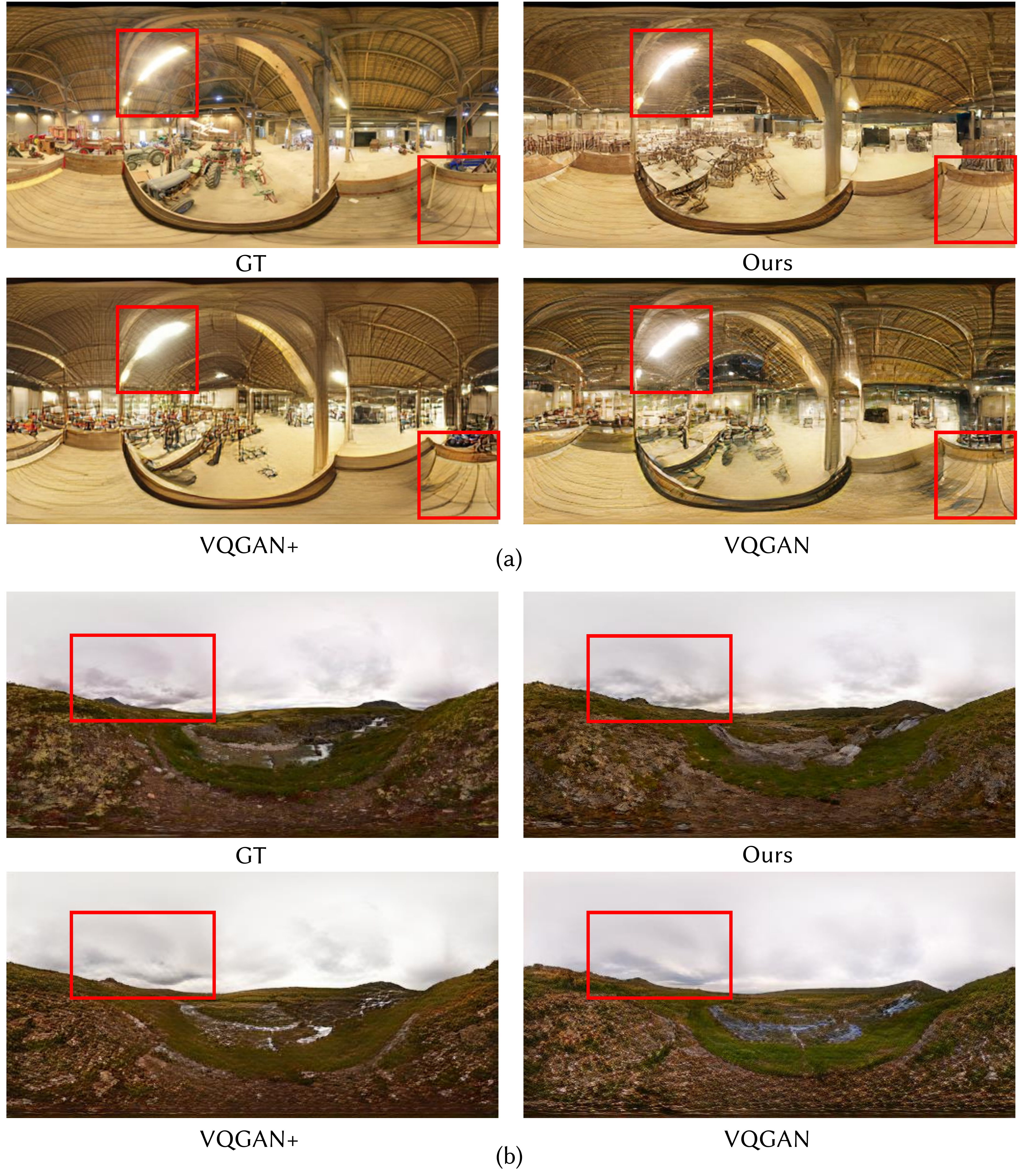}
  \caption{Qualitative comparison of panorama reconstruction with vanilla VQGAN. Especially, VQGAN+ denotes that VQGAN trained with the WS-perceptual loss~\cite{Akimoto2022DiverseP3}}
  \label{fig:supp_vqgan1}
\end{figure*}

\begin{figure*}[ht]
  \centering
  \includegraphics[width=0.7\linewidth]
  {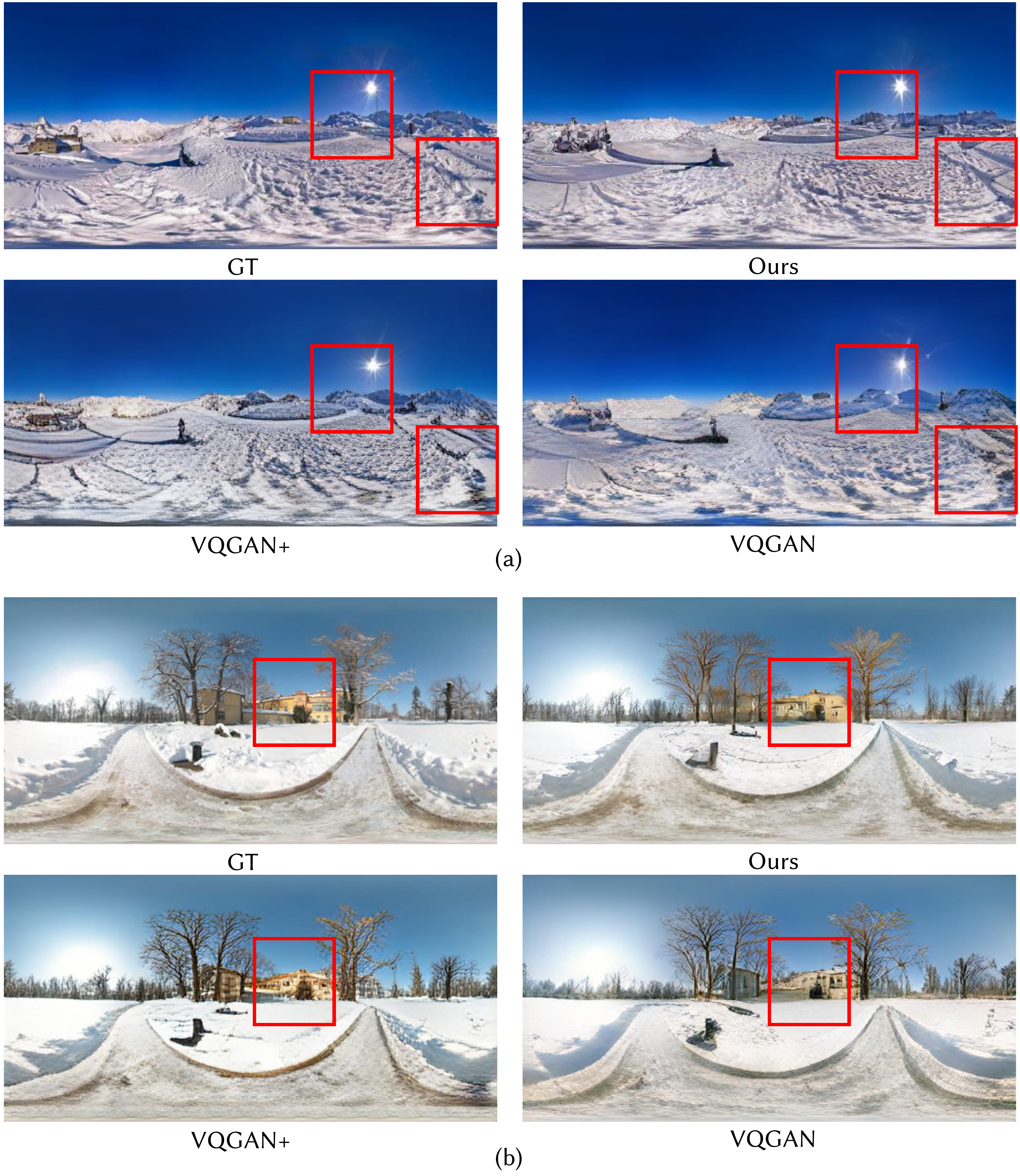}
  \caption{Qualitative comparison of panorama reconstruction with vanilla VQGAN. Especially, VQGAN+ denotes that VQGAN trained with the WS-perceptual loss~\cite{Akimoto2022DiverseP3}}
  \label{fig:supp_vqgan2}
\end{figure*}

\begin{figure*}[ht]
  \centering
  \includegraphics[width=0.7\linewidth]
  {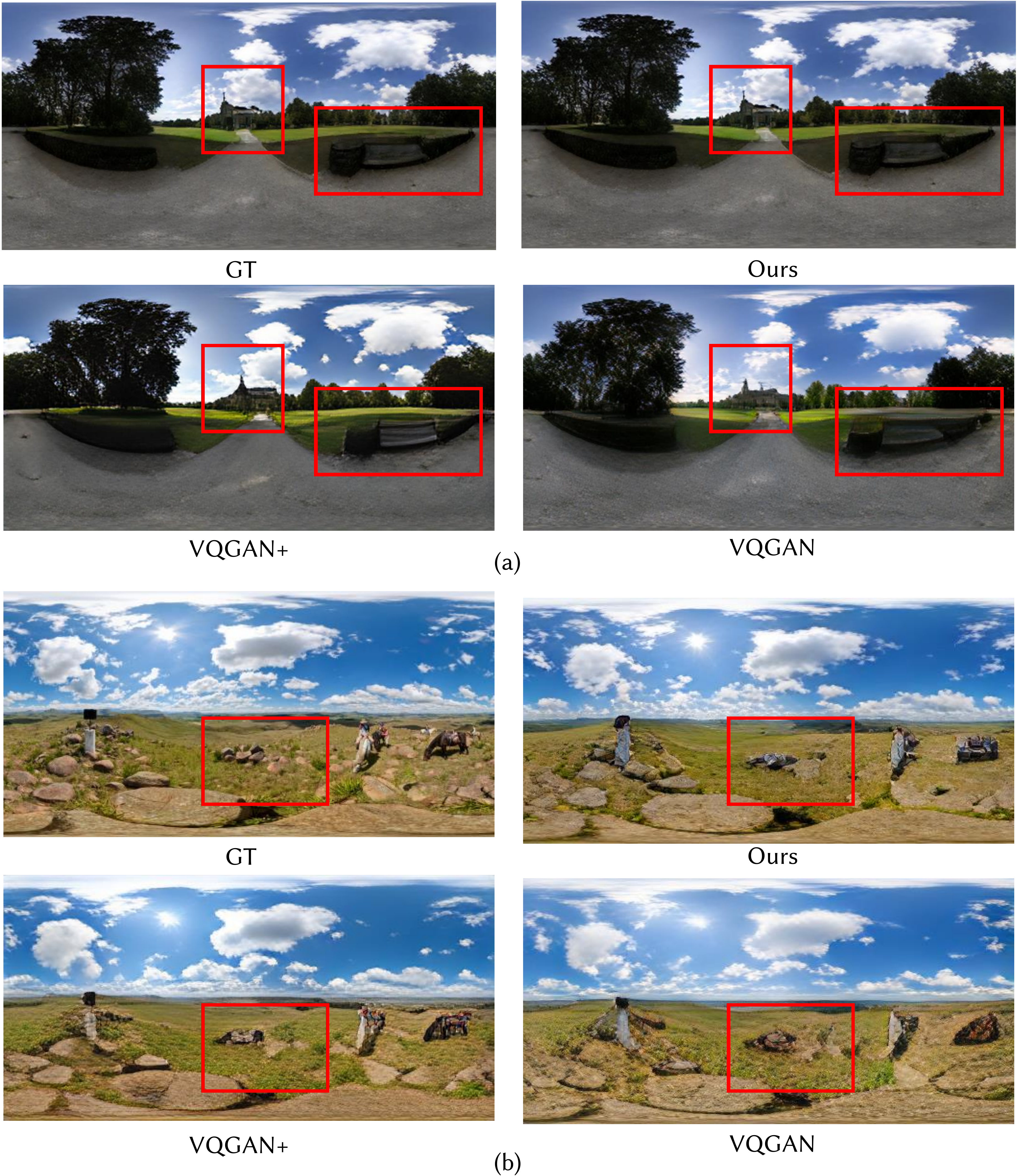}
  \caption{Qualitative comparison of panorama reconstruction with vanilla VQGAN. Especially, VQGAN+ denotes that VQGAN trained with the WS-perceptual loss~\cite{Akimoto2022DiverseP3}}
  \label{fig:supp_vqgan3}
\end{figure*}

In Fig.~\ref{fig:supp_vqgan1}, Fig.~\ref{fig:supp_vqgan2} and Fig.~\ref{fig:supp_vqgan3}, we compare the reconstruction results between our S-VQGAN, vanilla VQGAN and VQGAN+. Note that VQGAN+ represents VQGAN with WS-perceptual loss~\cite{Akimoto2022DiverseP3} and our S-VQGAN is only trained with the conventional loss functions. As the figures show, our S-VQGAN reconstructs more smooth images and better structural details than  VQGAN and VQGAN+, which further verifies the capability of the sphere-specific codebook.
\begin{table}[t]
\centering
  \caption{Reconstruction quality of complete panoramas with different resolution sizes. The best performance is highlighted in red. }
  \label{tab:diff_result}
  \resizebox{0.8\linewidth}{!}{ \begin{tabular}{cccc}
    \toprule
   Resolution &  Method  &WS-PSNR$\uparrow$&FID$\downarrow$\\
    \midrule
    \multirow{3}{*}{256$\times$512 }&VQGAN &  13.11& 16.22\\
    & VQ-GAN+& 13.41& 13.83\\
    & S-VQGAN&\cellcolor{red!30}13.72&\cellcolor{red!30}12.64\\
  \bottomrule
\end{tabular}}
\end{table}
\begin{figure*}[ht]
  \centering
  \includegraphics[width=\linewidth]
  {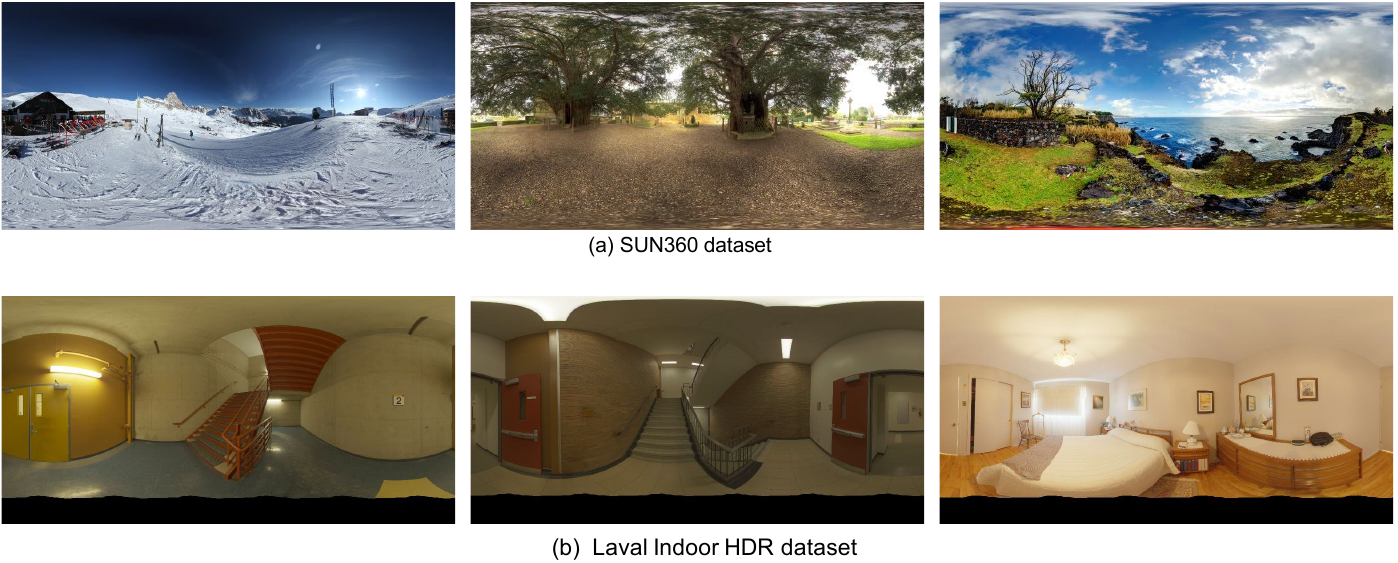}
    \vspace{-10pt}
  \caption{The visualization of the dataset.}
  \vspace{-10pt}
  \label{fig:sup_dataset}
\end{figure*}

\subsection{Diverse Panorama Outpainting}
\begin{figure*}[h]
  \centering
  \includegraphics[width=0.65\linewidth]
  {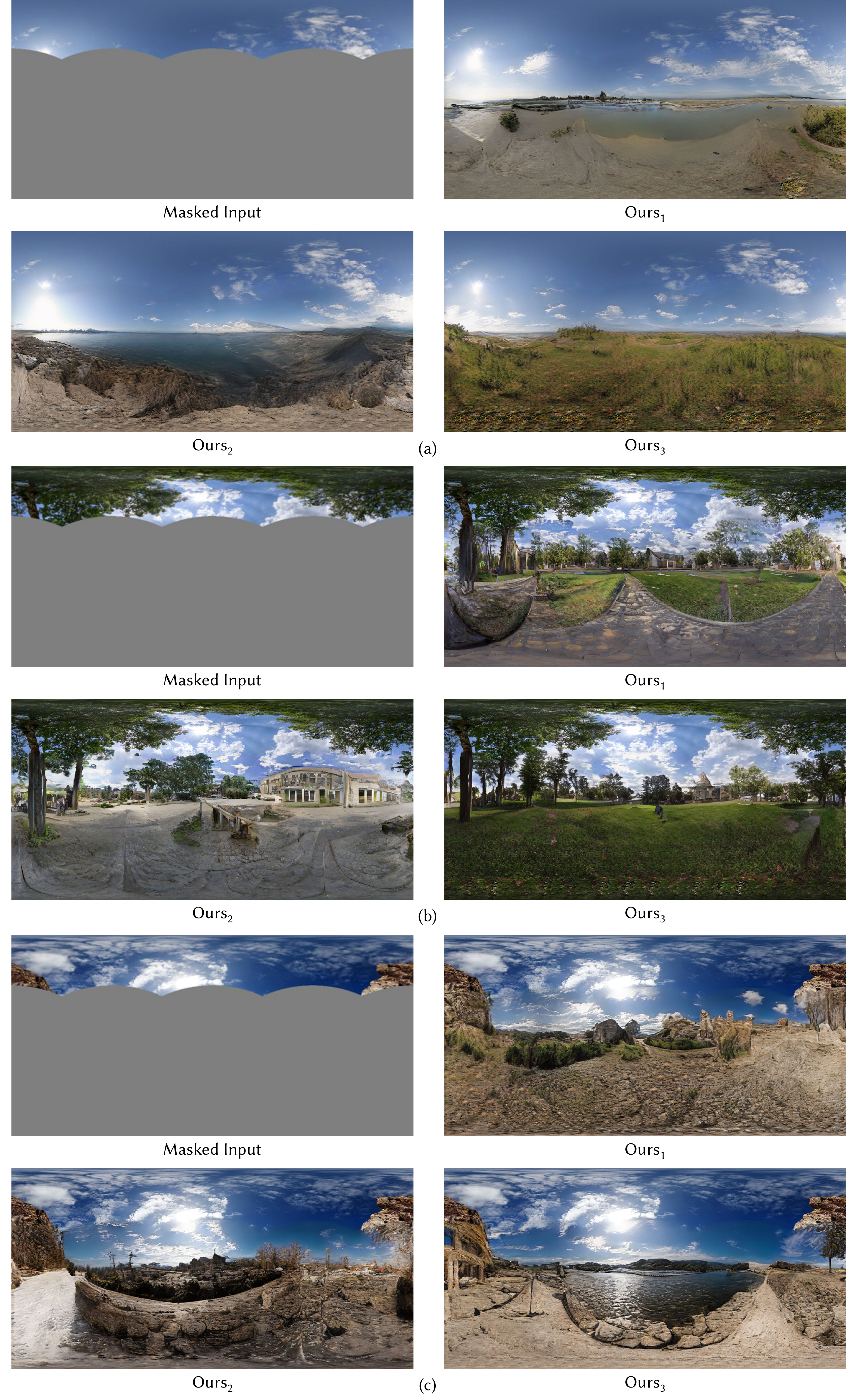}
  \caption{Qualitative results on SUN360 dataset. The masked inputs are at the ceil.}
  \label{fig:supp_results3}
\end{figure*}

\begin{figure*}[h]
  \centering
  \includegraphics[width=0.65\linewidth]
  {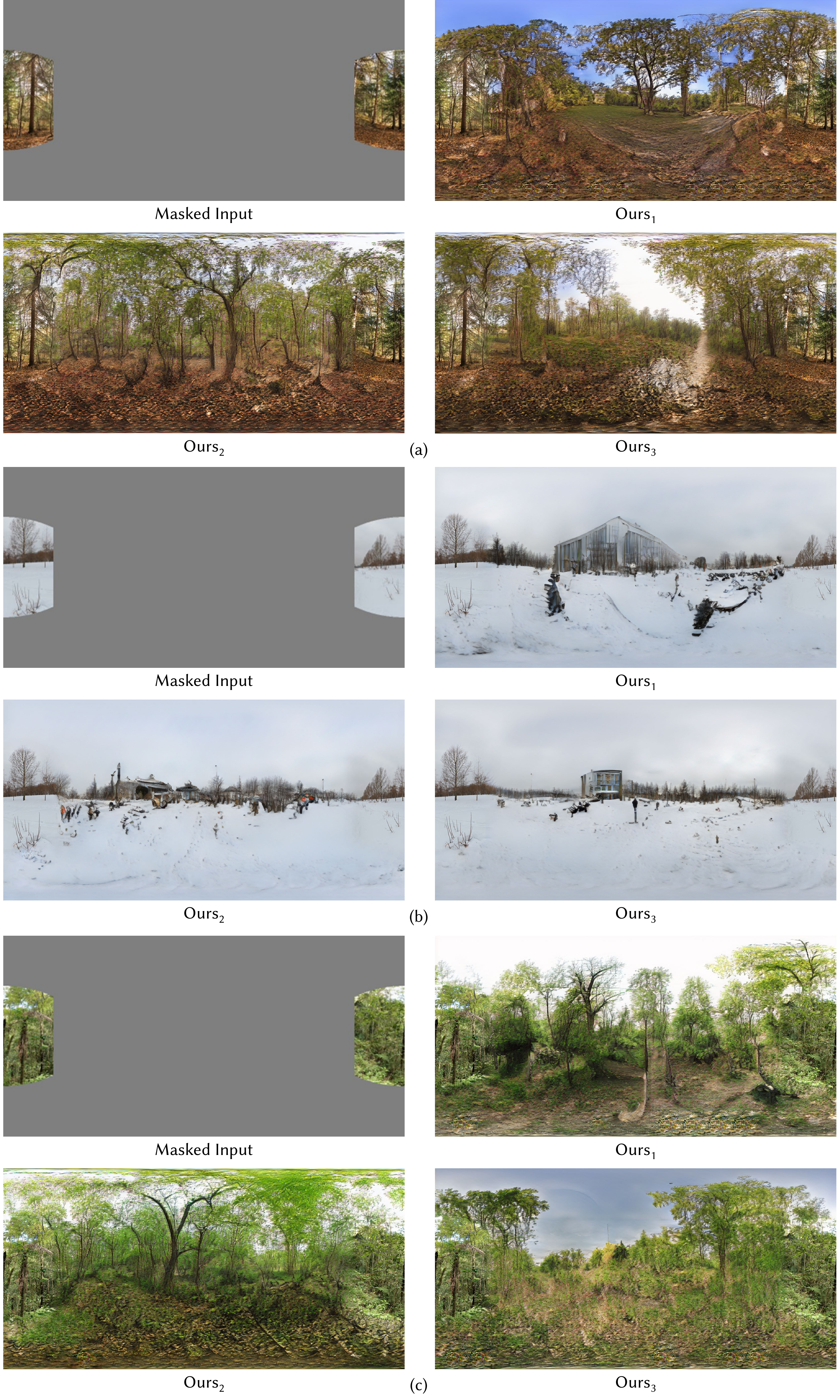}
  \caption{Qualitative results on SUN360 dataset. The masked inputs are at the left.}
  \label{fig:supp_results4}
\end{figure*}

\begin{figure*}[h]
  \centering
  \includegraphics[width=0.65\linewidth]
  {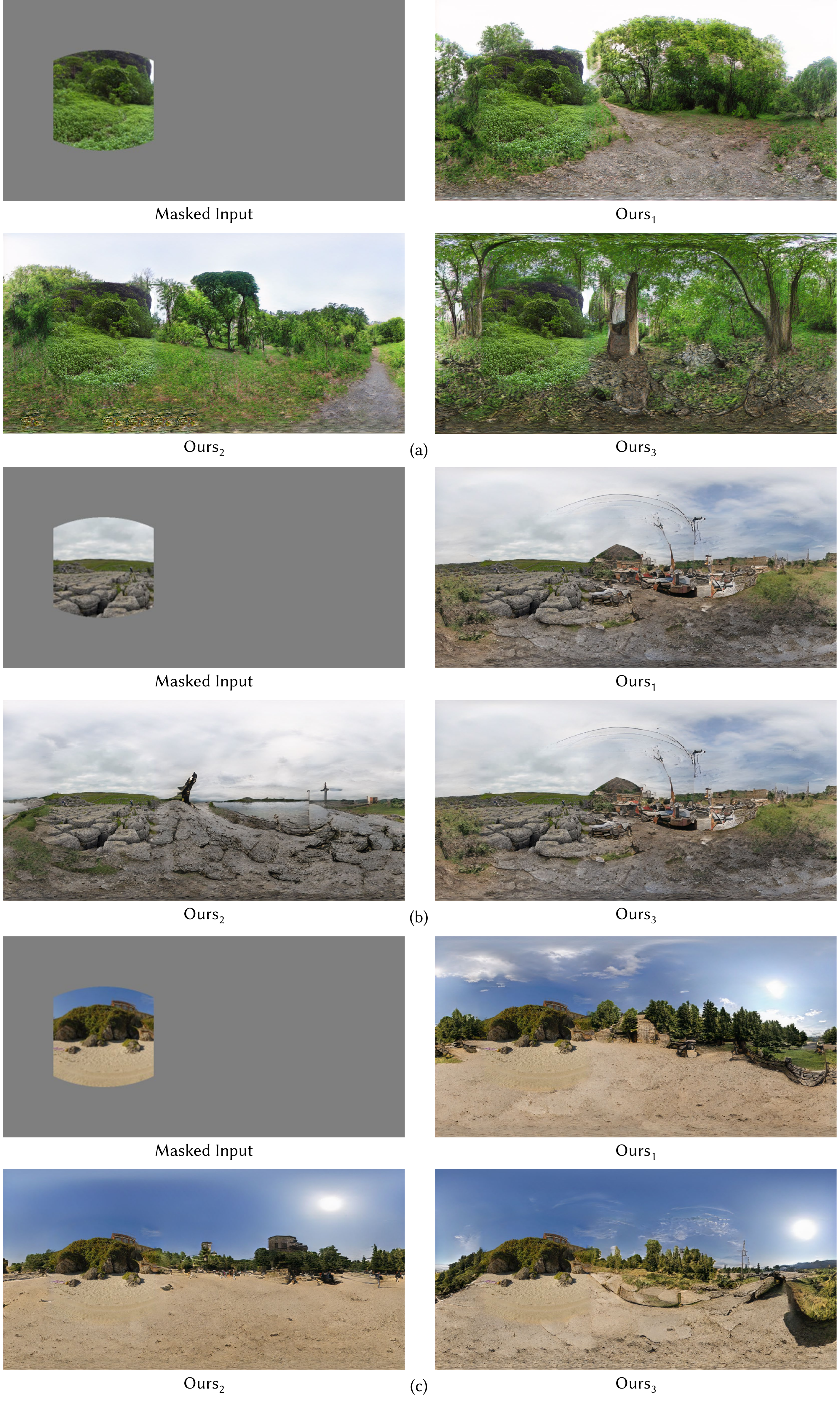}
  \caption{Qualitative results on SUN360 dataset. The masked inputs are at the front.}
  \label{fig:supp_results6}
\end{figure*}

\begin{figure*}[h]
  \centering
  \includegraphics[width=0.65\linewidth]
  {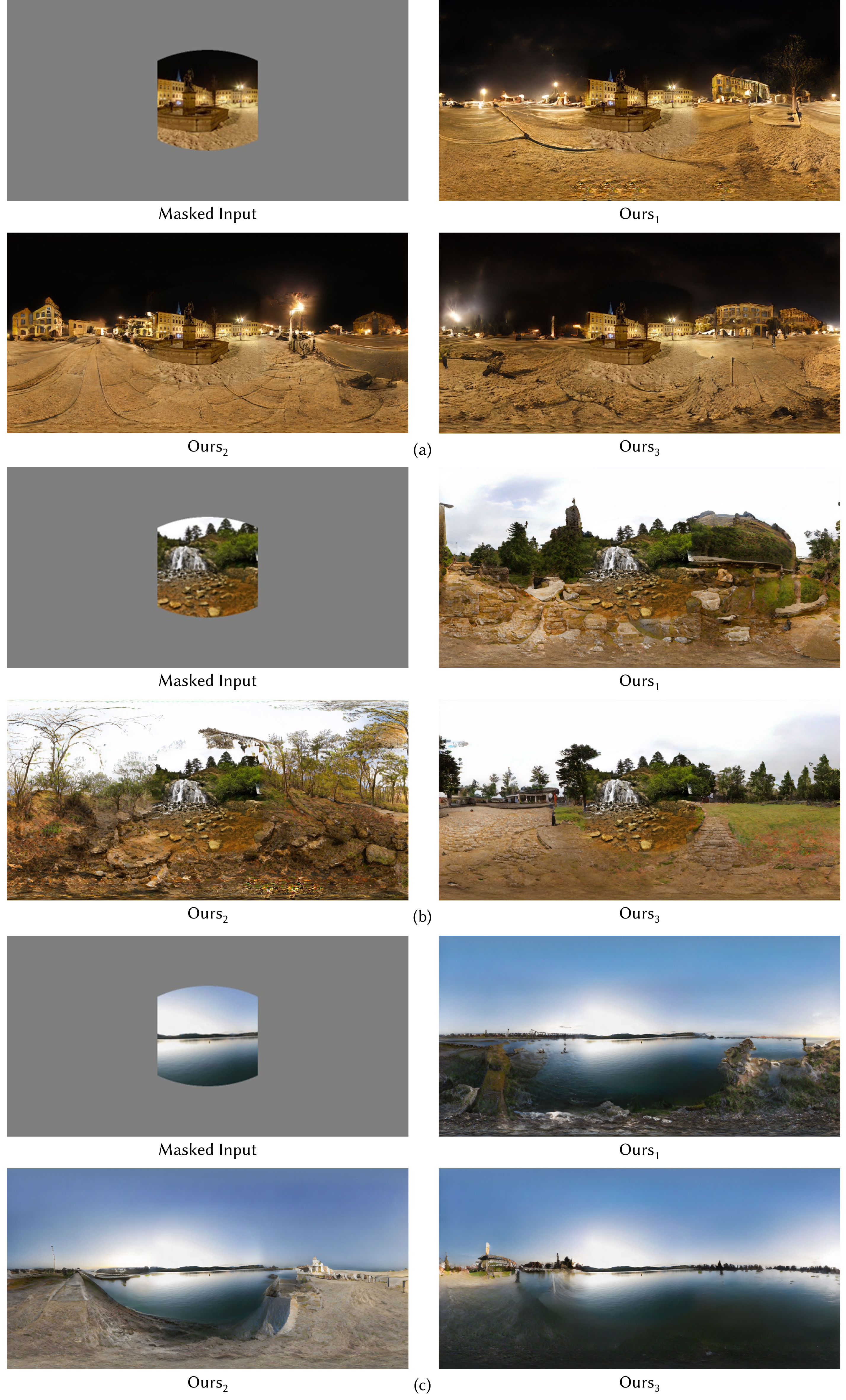}
  \caption{Qualitative results on SUN360 dataset. The masked inputs are at the right.}
  \label{fig:supp_results5}
\end{figure*}

\begin{figure*}[h]
  \centering
  \includegraphics[width=0.65\linewidth]
  {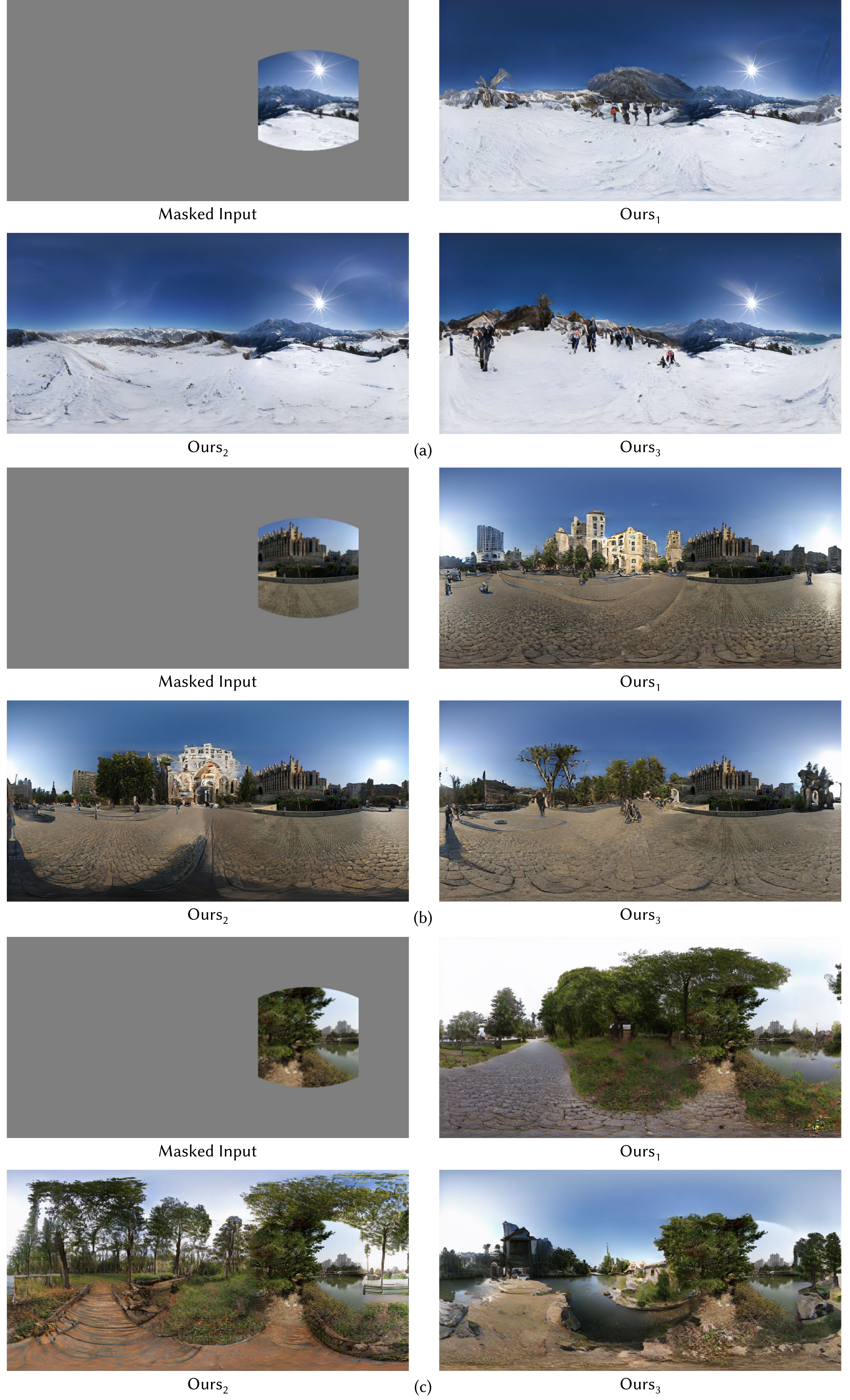}
  \caption{Qualitative results on SUN360 dataset. The masked inputs are at the back.}
  \label{fig:supp_results2}
\end{figure*}

\begin{figure*}[h]
  \centering
  \includegraphics[width=0.65\linewidth]
  {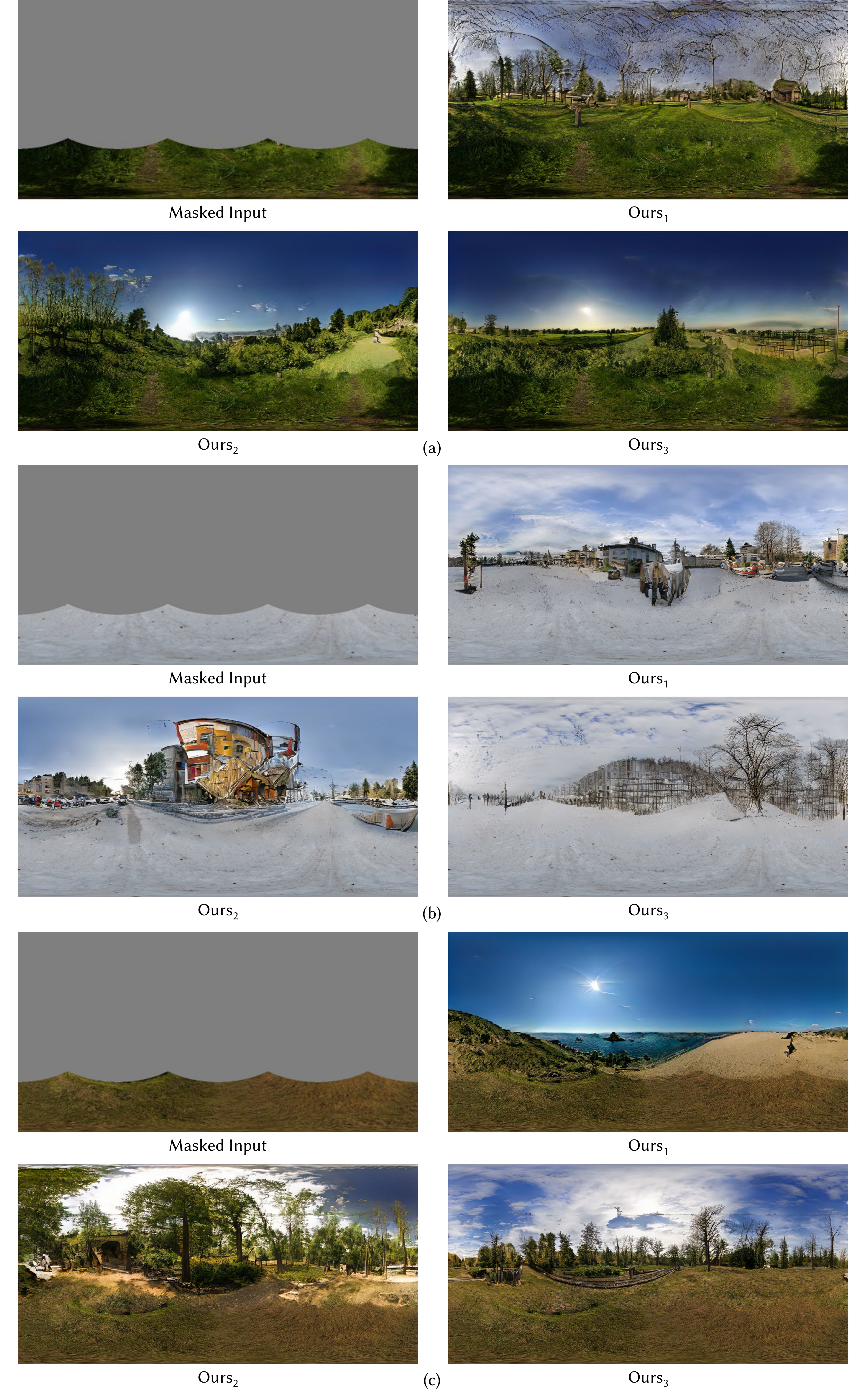}
  \caption{Qualitative results on SUN360 dataset. The masked inputs are at the floor.}
  \label{fig:supp_results1}
\end{figure*}

In Fig.~\ref{fig:supp_results3}, Fig.~\ref{fig:supp_results4}, 
Fig.~\ref{fig:supp_results6},
Fig.~\ref{fig:supp_results5},
Fig.~\ref{fig:supp_results2},  
and Fig.~\ref{fig:supp_results1}, we show our diverse high-fidelity and high-resolution outpainting results at different locations on SUN360 dataset. 
The extensive visual results show that our Dream360 can handle the flexible inputs and obtain the visually realistic outpainting results.

\subsection{Qualitative Comparisons to Existing SoTA Methods}

\begin{figure*}[h]
  \centering
  \includegraphics[width=0.65\linewidth]
  {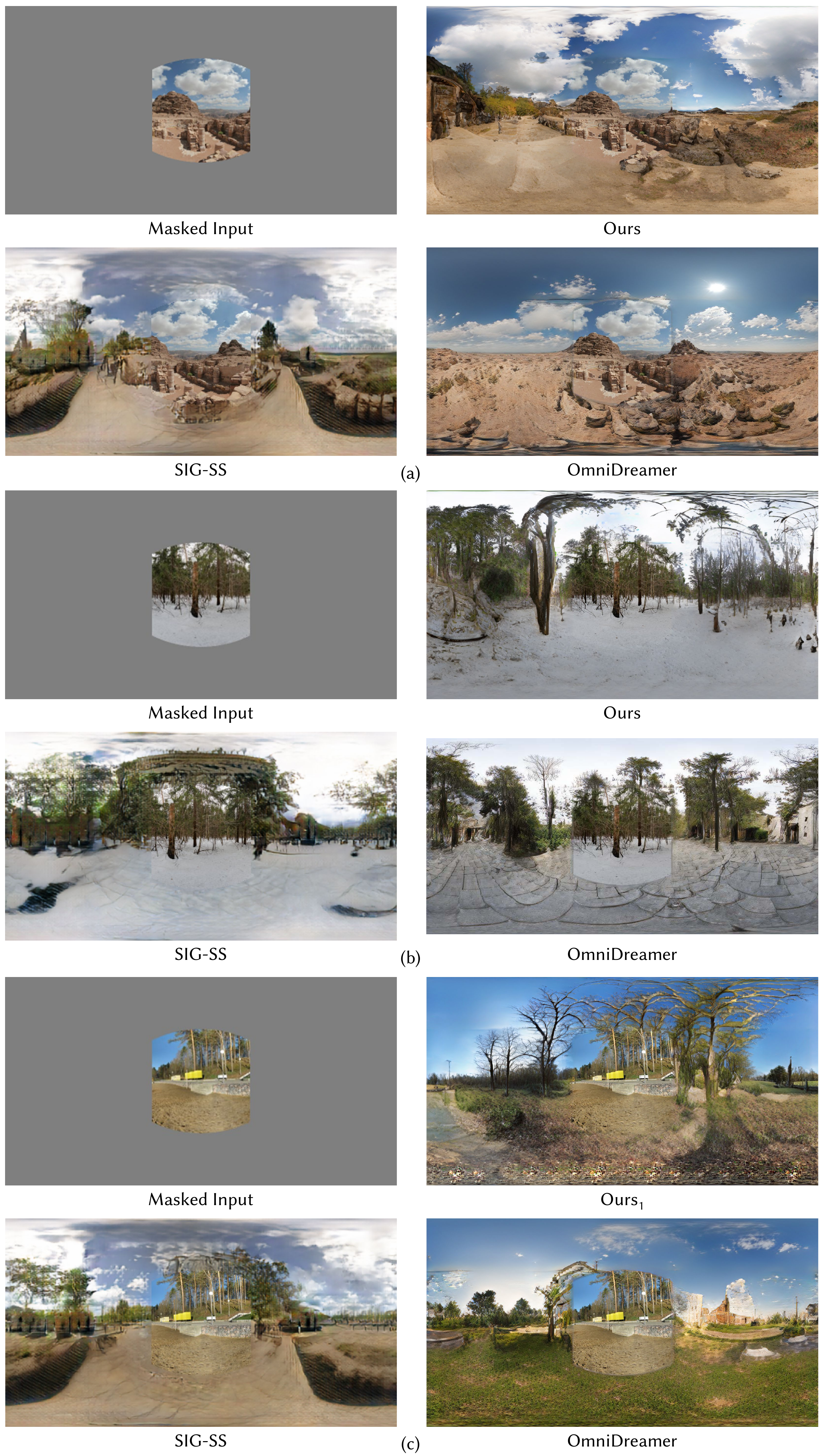}
  \caption{Qualitative comparison with SIG-SS~\cite{Hara2021SphericalIG} and Omnidreamer~\cite{yu2021diverse} on SUN360 dataset. We can see that the outpainting results of our Dream360 are more visually realistic and contains more structural details.}
  \label{fig:supp_comp1}
\end{figure*}

\begin{figure*}[h]
  \centering
  \includegraphics[width=0.65\linewidth]
  {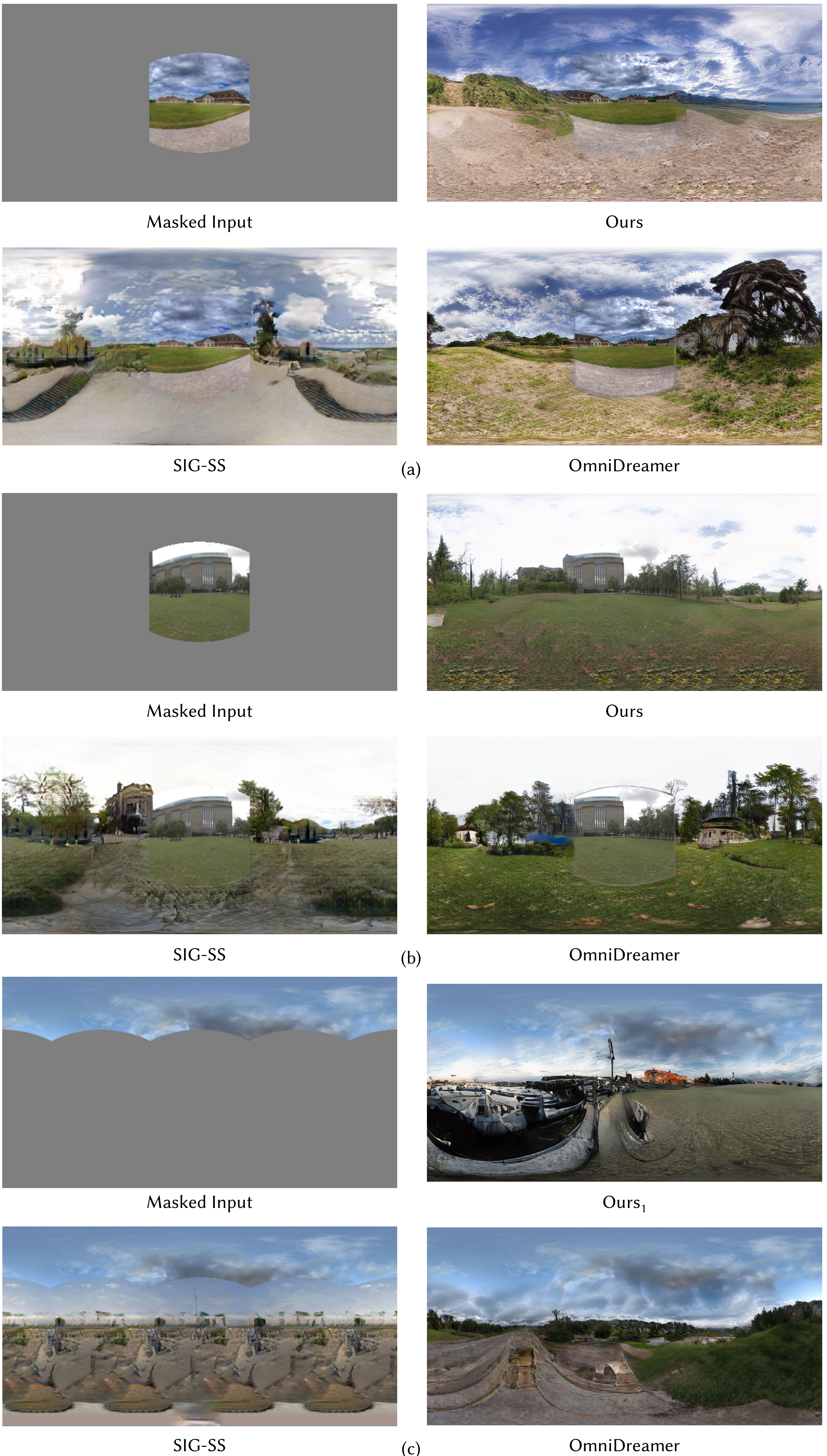}
  \caption{Qualitative comparison with SIG-SS~\cite{Hara2021SphericalIG} and Omnidreamer~\cite{yu2021diverse} on SUN360 dataset. From the results, we can observe that our Dream360 can handle the difficult condition as SIG-SS failed to restore the content from the input with the ceil mask. }
  \label{fig:supp_comp2}
\end{figure*}

In Fig.~\ref{fig:supp_comp1} and Fig.~\ref{fig:supp_comp2},
we provide more comparison results with existing SoTA methods and verify the superior of Dream360.
\subsection{Ablation Studies}
\begin{figure*}[h]
  \centering
  \includegraphics[width=0.65\linewidth]
  {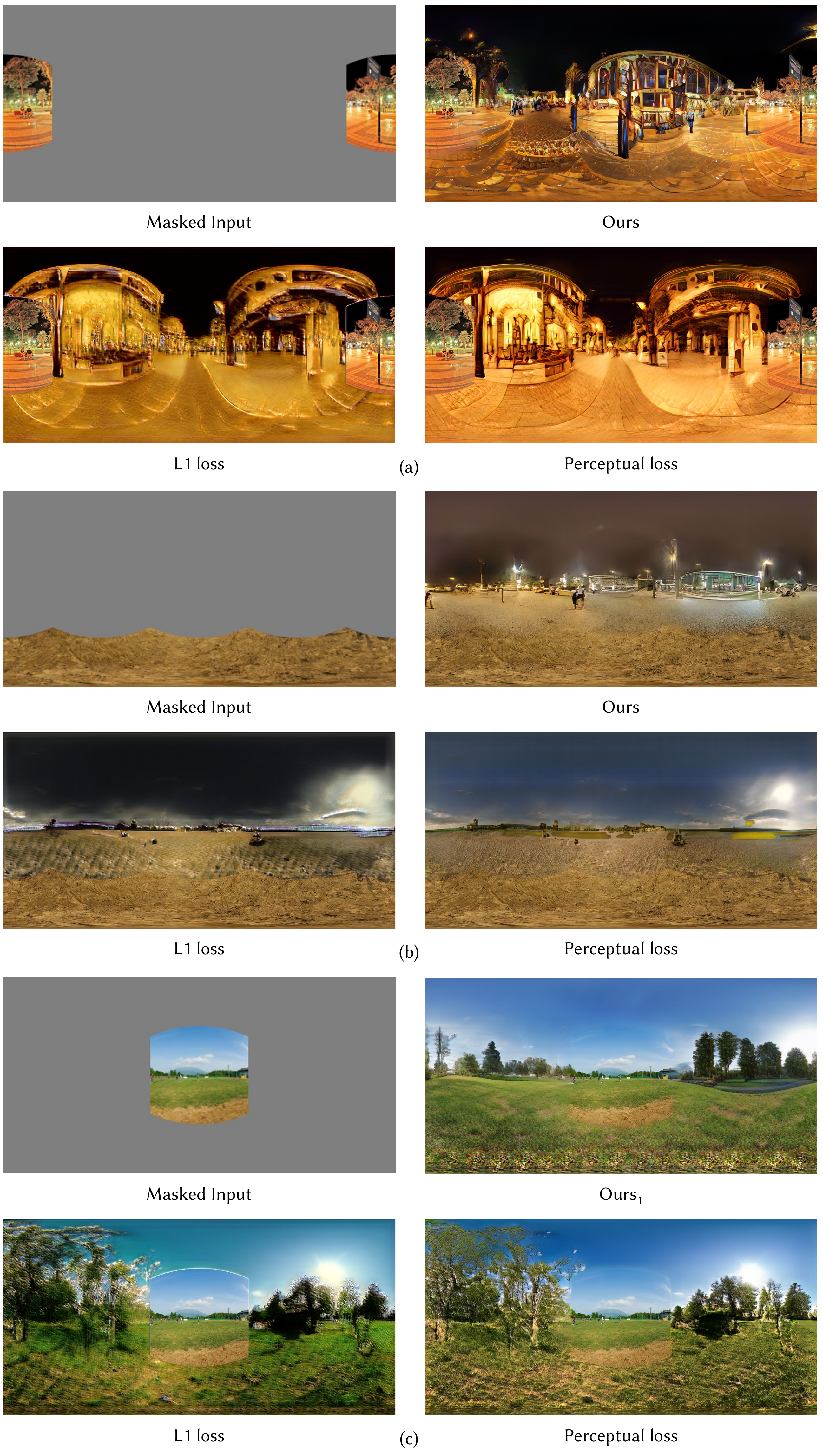}
  \caption{Qualitative comparison with SIG-SS~\cite{Hara2021SphericalIG} and Omnidreamer~\cite{yu2021diverse}. We can see that the outpainting results of our Dream360 are more visually realistic and contains more structural details.}
  \label{fig:ab1}
\end{figure*}

\begin{figure*}[h]
  \centering
  \includegraphics[width=0.65\linewidth]
  {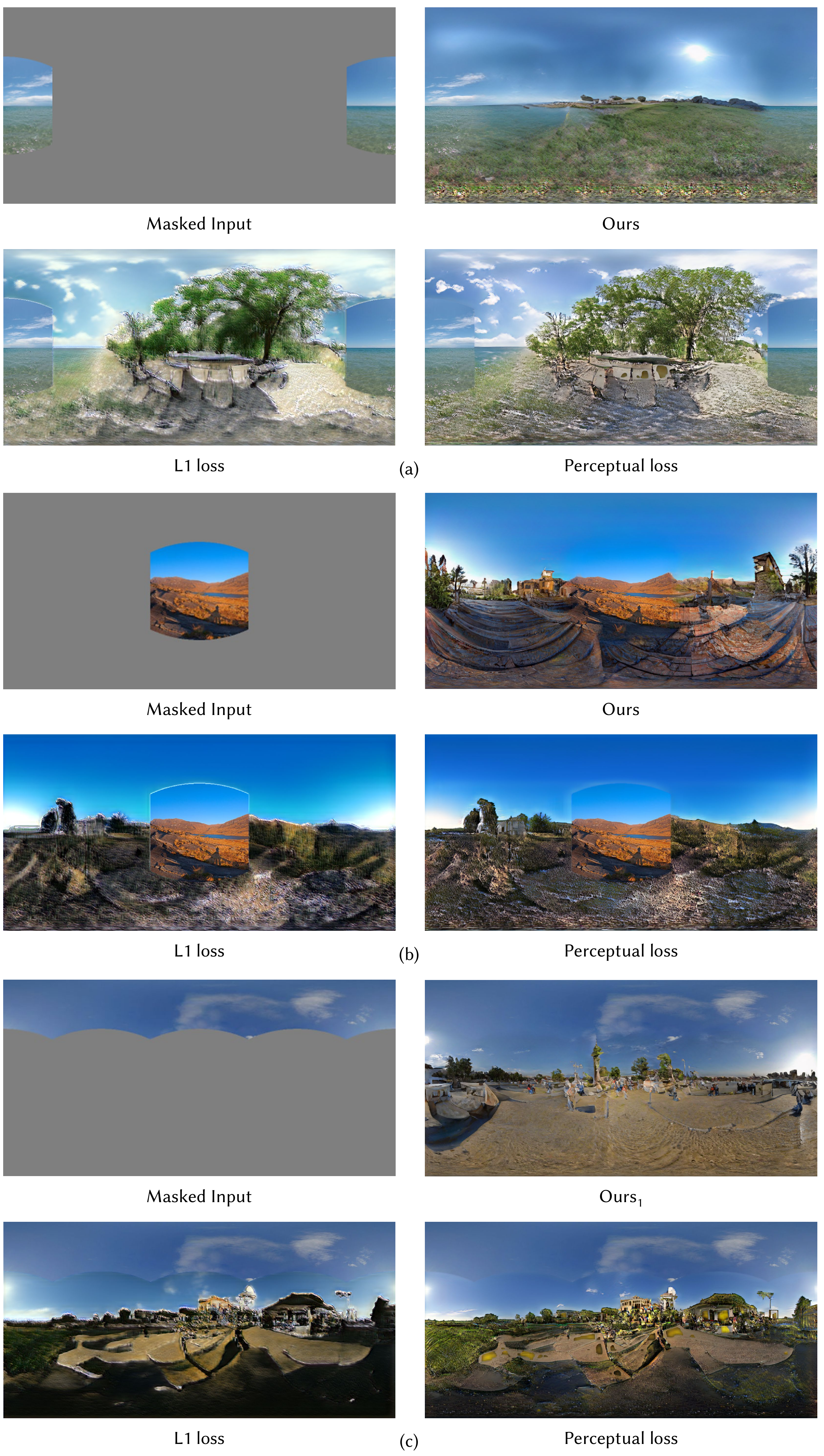}
  \caption{Qualitative comparison with SIG-SS~\cite{Hara2021SphericalIG} and Omnidreamer~\cite{yu2021diverse}. From the results, we can observe that our Dream360 can handle the difficult condition as SIG-SS failed to restore the content from the input with the ceil mask. }
  \label{fig:ab2}
\end{figure*}

Fig.~\ref{fig:ab1} and Fig.~\ref{fig:ab2} show more comparison results between L1 loss in the frequency domain, perceptual loss in the frequency domain and our frequency-aware consistency loss. The results show that our loss can further enhance the high-frequency component and improve the visual fidelity.


\section{Experimental Results on Indoor Scene Creation}
\subsection{Dataset}
To further eveluate our Dream360, we conduct extensive experiments on the Laval Indoor dataset. Regarding the Laval Indoor dataset, as shown in Fig.~\ref{fig:sup_dataset}(b), which consists of 2,100 HDR indoor panoramas, we followed the train/test split ratio as described in ~\cite{Gardner2017LearningTP}, using 1,719 panoramas for training and 289 panoramas for testing. Since the Laval Indoor dataset is relatively small, we first trained the Stage $\RN{1}$ model on the SUN360 dataset and subsequently fine-tuned it using the Laval Indoor dataset. For Stage $\RN{2}$, we directly employed the model trained on the SUN360 dataset. It is important to note that the Laval Indoor dataset exhibits a black region at the bottom. Consequently, when conducting the outpainting experiment, we excluded the ceiling and floor regions from our masked input.
\subsection{Qualitative results on Laval Indoor dataset}
\begin{figure*}[h]
  \centering
  \includegraphics[width=0.6\linewidth]
  {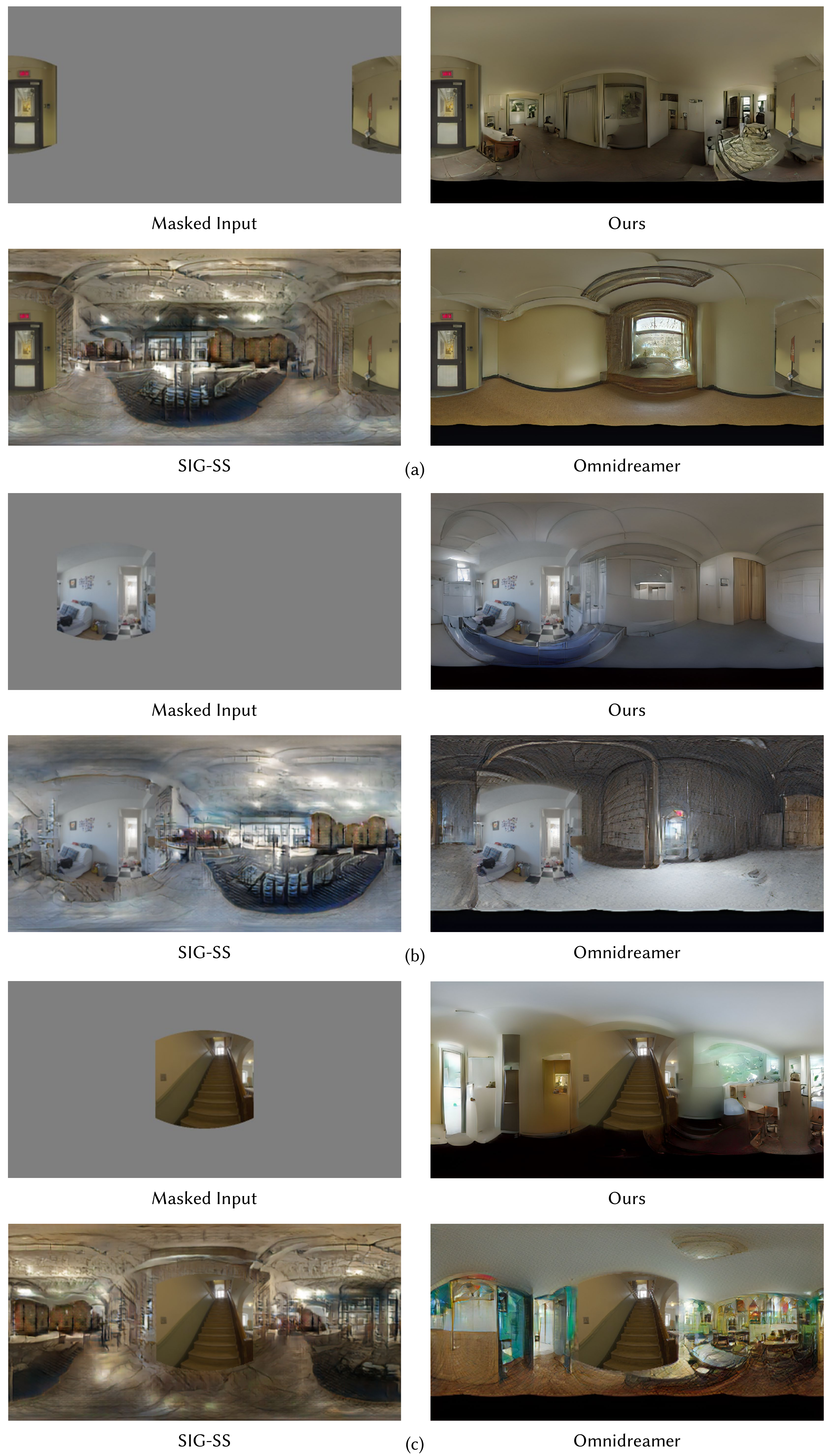}
  \caption{Qualitative comparison with SIG-SS~\cite{Hara2021SphericalIG} and Omnidreamer~\cite{yu2021diverse} on Laval Indoor dataset.}
  \label{fig:supp_comp3}
\end{figure*}

\begin{figure*}[h]
  \centering
  \includegraphics[width=0.65\linewidth]
  {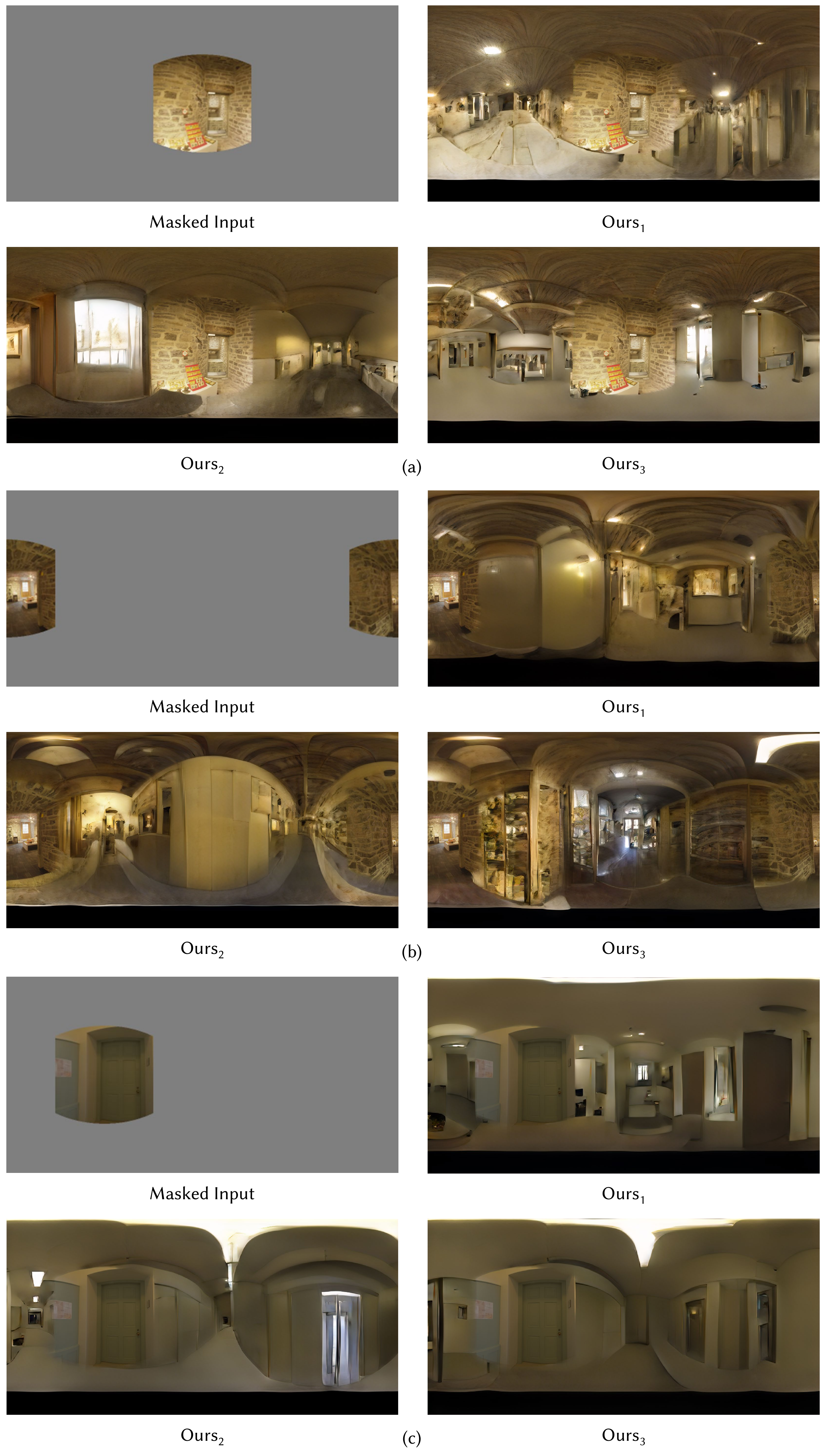}
  \caption{Qualitative results on Laval Indoor dataset.}
  \label{fig:supp_results7}
\end{figure*}

In Fig.~\ref{fig:supp_comp3}, we provide the comparison results with existing SoTA methods on Laval Indoor dataset and verify the superior of Dream360.
In Fig.~\ref{fig:supp_results7} shows the outpainting results on Laval Indoor dataset. The extensive visual results show that our Dream360 can handle the flexible inputs even for the indoor scenes and obtain the visually realistic outpainting results.

\section{Experimental results on Real-world Data}
\begin{figure*}[h]
  \centering
  \includegraphics[width=0.65\linewidth]
  {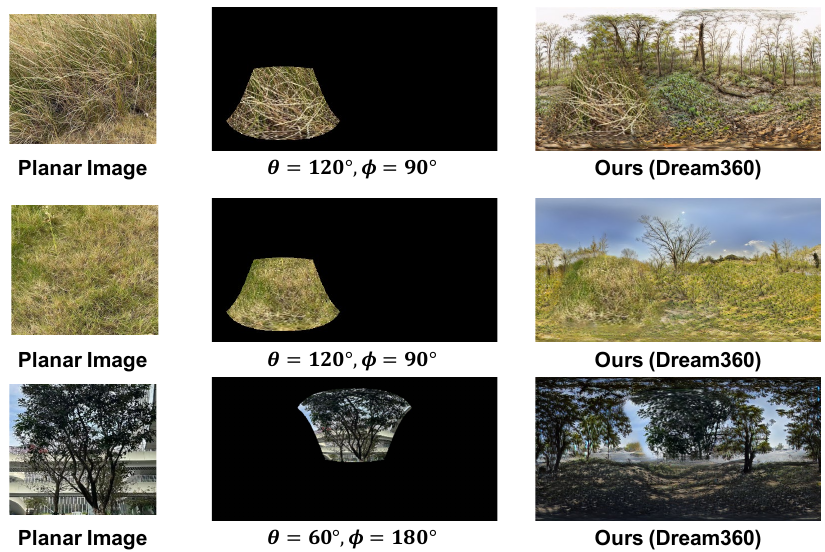}
  \caption{Qualitative results with the input taken by participants.}
  \label{fig:exten_self}
\end{figure*}

As shown in Fig.~\ref{fig:exten_self}, 
we generate complete panoramas using planar images captured by the participants. Specifically, the participant provides planar images taken with a smartphone or pinhole camera. We utilize the geometric relationships between CP and ERP to convert the planar images into masked panoramas based on viewpoint directions selected by the participant. Subsequently, we use the Dream360 model pre-trained on the SUN360 dataset to generate complete panoramas from these masked input.

\section{Failure Cases}
\begin{figure*}[h]
  \centering
  \includegraphics[width=0.8\linewidth]
  {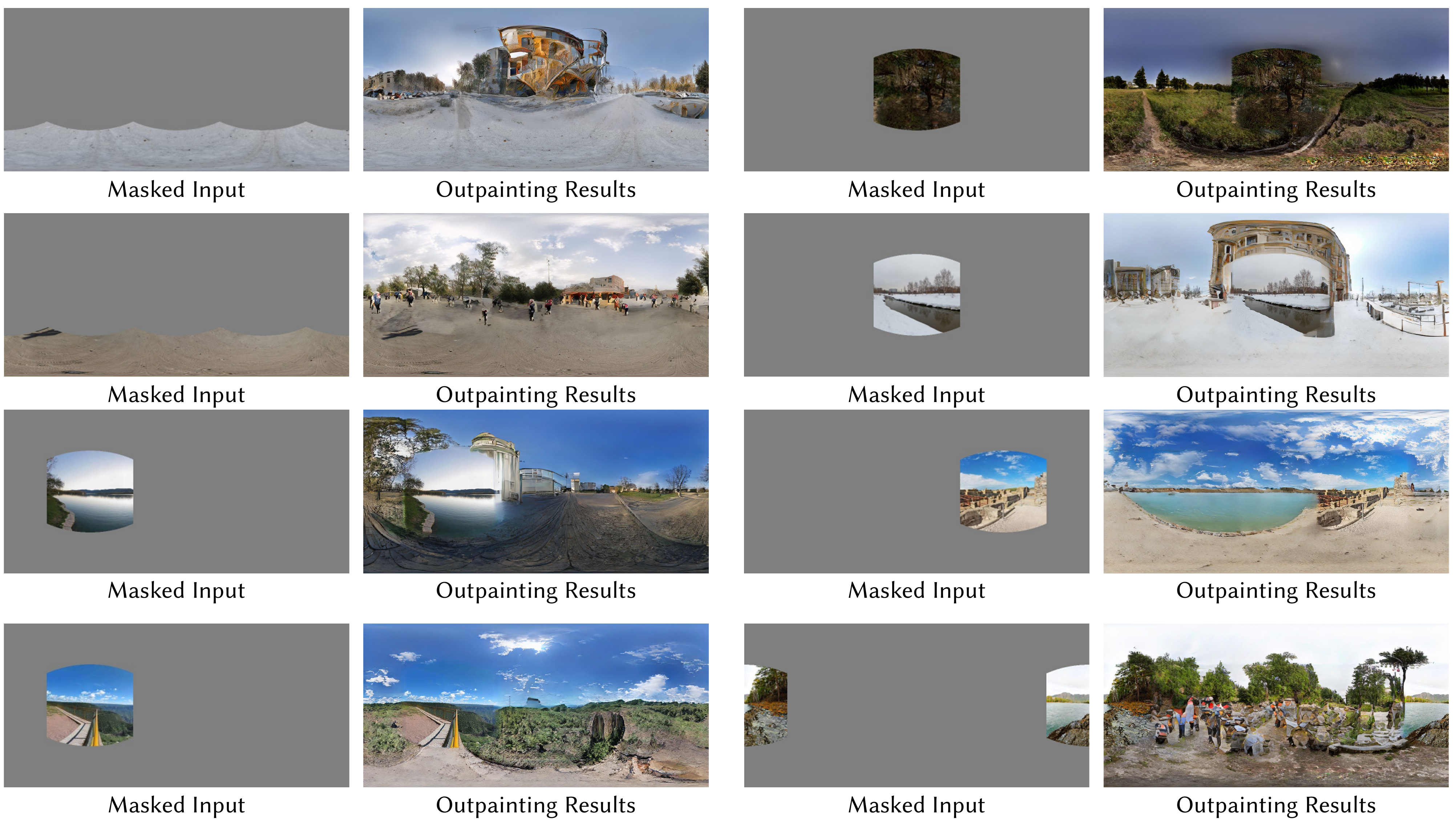}
  \caption{Failure cases.}
  \label{fig:fail}
\end{figure*}

As Fig.~\ref{fig:fail} shows, although our model tends to restore more structural details, it is weak in generating complex objects, \eg human and large building. Besides, sometimes the generated content is inconsistent with the given input. In our future work, we target to generate more precise objects and reduce the content inconsistency between the given masked input and generated region.

\clearpage
\bibliographystyle{abbrv-doi}

\bibliography{template}